\documentclass[twoside,11pt]{article}

\usepackage{jmlr2e}
\usepackage{bm}
\usepackage{bbm}
\usepackage{amssymb}
\usepackage{amsmath}
\usepackage{float}
\usepackage{comment}
\usepackage{color}
\usepackage{authblk}
\firstpageno{1}

\title{Minimizing Negative Transfer of Knowledge in Multivariate Gaussian Processes: A Scalable and Regularized Approach}

\author[1]{Raed Kontar}
\author[2]{Garvesh Raskutti}
\author[3]{Shiyu Zhou}

\affil[1]{Department of Industrial \& Operations Engineering, University of Michigan}
\affil[2]{Department of Statistics, University of Wisconsin-Madison}
\affil[3]{Department of Industrial \& Systems Engineering, University of Wisconsin-Madison}
\date{}                     %% if you don't need date to appear
\begin{document}

\maketitle

\vspace{1em}

\begin{abstract}%   <- trailing '%' for backward compatibility of .sty file
Recently there has been an increasing interest in the multivariate Gaussian process (MGP) which extends the Gaussian process (GP) to deal with multiple outputs. One approach to construct the MGP and account for non-trivial commonalities amongst outputs employs a convolution process (CP). The CP is based on the idea of sharing latent functions across several convolutions. Despite the elegance of the CP construction, it provides new challenges that need yet to be tackled. First, even with a moderate number of outputs, model building is extremely prohibitive due to the huge increase in computational demands and number of parameters to be estimated. Second, the negative transfer of knowledge may occur when some outputs do not share  commonalities. In this paper we address these issues.  We propose a regularized pairwise modeling approach for the MGP established using CP. The key feature of our approach is to distribute the estimation of the full multivariate model into a group of bivariate GPs which are individually built.  Interestingly pairwise modeling turns out to possess unique characteristics, which allows us to tackle the challenge of negative transfer through penalizing the latent function that facilitates information sharing in each bivariate model.  Predictions are then made through combining predictions from the bivariate models within a Bayesian framework. The proposed method has excellent scalability when the number of outputs is large and minimizes the negative transfer of knowledge between uncorrelated outputs. Statistical guarantees for the proposed method are studied and its advantageous features are demonstrated through numerical studies.
\end{abstract}

\begin{comment} 
Recently there has been an increasing interest in the multivariate Gaussian process (MGP) which extends the Gaussian process (GP) to deal with multiple outputs. One approach to construct the MGP and account for non-trivial commonalities amongst output employs a convolution process (CP). The CP is based on the idea of sharing latent functions across several convolutions. Despite the elegance of the CP construction, it provides new challenges that need yet to be tackled. First, even with a moderate number of outputs, model building is extremely prohibitive due to the huge increase in computational demands and number of parameters to be estimated. Second, the negative transfer of knowledge may occur when some outputs do not share some commonalities.  In this paper we address these issues. We propose a regularized pairwise modeling approach for MGP models established using CP. The key feature of our approach is to decompose the full multivariate model into a group of bivariate GP where the shared latent function parameters that facilitate the sharing of information in each bivariate model are penalized. Predictions are then made through combining predictions from the bivariate models within a Bayesian framework. The proposed method has excellent scalability when the number of outputs is large and minimizes the negative transfer of knowledge between uncorrelated outputs. Statistical guarantees for the proposed method are studied and its advantageous features are demonstrated through extensive numerical studies.
\end{comment}


\vspace{3em}

\section{Introduction}

Gaussian process regression models are widely used in several fields due to their desirable properties, such as flexibility, ease of implementation, uncertainty quantification and natural Bayesian interpretation~\citep{quinonero2005unifying}. Recently, there has been increasing interest in extending GP models to deal with multivariate outputs (also known as \emph{cokriging}) due to their prevalence in many applications. For example, in manufacturing plants, hard to sample performance indicators can be predicted from other correlated and cheap to sample indicators~\citep{osborne2008towards}. Also, the future evolution of sensor signals from in-service devices can be predicted using previously observed signals from similar devices in the historical database
~\citep{kontar2017nonparametrica}. Other applications arise in geostatistics, wireless networks and computer experiments.

The multivariate Gaussian process draws its roots from multitask learning. When multiple datasets from related outputs exist, integrative analysis can be advantageous relative to learning outputs independently. This integrative analysis, which leverages commonalities among related outputs to improve prediction and learning accuracy is referred to as multitask learning ~\citep{caruana1998multitask, yuan2012visual}.  The key feature in multitask learning is to provide a shared representation between training and testing outputs to allow the inductive transfer of knowledge. From an MGP perspective, this transfer of knowledge is achieved through specifying a valid positive semidefinite covariance function that models the dependencies of all data points within an output and across different outputs~\citep{conti2009gaussian}.

Traditionally, in MGP models, outputs are jointly modeled with a separable covariance structure, i.e. 
correlation over the same the input space is separable from between-output correlation~\citep{conti2010Bayesian,qian2008gaussian,zhou2011simple}. In such methods, the separable covariance function is of the form $t \times \mbox{cov}(\bm{x},\bm{x}^{\prime})$ where $t$ is the between-output covariance matrix and $\mbox{cov}(.,.)$ is a covariance function over inputs $\bm{x}\in \mathcal{R}^D$, the same for all outputs. This assumption is appealing due to the simplified covariance structure and significant reduction of model parameters, however it restricts all marginal GP's for all outputs to share the same set of covariance parameters defined in $\mbox{cov}(.,.)$. 

On the other hand, nonseparable covariance functions allow outputs to possess both shared and unique features, as different outputs share information through different covariance parameters, therefore, accounting for non-trivial commonalities in the data. Recent work on nonseparable covariance functions are mainly based on convolution processes (CP)~\citep{majumdar2007multivariate,fricker2013multivariate,melkumyan2011multi}. Earlier work, is known in the geostatistics literature as the linear model of coregionalization (LMC)~\citep{goulard1992linear} and can be seen as a special case of the CP framework~\citep{alvarez2010efficient}. The CP is based on the idea that a GP can be constructed by convolving a latent function with a smoothing kernel~\citep{ver1998constructing}. Thus, if each output is expressed as a convolution of a latent function drawn from a GP, and if we share these latent functions across multiple convolutions, then, multiple outputs can be expressed as a jointly distributed GP~\citep{boyle2005dependent,alvarez2009sparse}. As referred to by~\cite{alvarez2011computationally}, the key feature of the CP approach is facilitating the non-instantaneous mixing of base processes, where, for instance, each output can be described using its own length-scale,  which gives an added flexibility for describing the data.

Despite the elegance of the MGP established using CP, denoted as MGCP, model building is extremely prohibitive and often impractical even with a moderate number of outputs. Further, it is not uncommon for modern engineering systems to have a large number of outputs~\citep{higdon2008computer, mcfarland2008calibration}. For instance, when considering qualitative factors, the number of outputs in computer experiments increases dramatically based on all potential combinations~\citep{han2009prediction}. This leads to a set of considerable challenges. 

\begin{itemize}
\item \emph{Challenge 1} (\emph{Computational complexity}): The fact that the full covariance function of the joint GP should be considered results in significant computational burden and numerical issues. This challenge, in fact, is a direct consequence of accounting for multiple outputs and has been recently tackled in some literature~\citep{alvarez2011computationally,kontar2017nonparametricb}.
\end{itemize}

\noindent
However, two other challenges arise with the CP construction and have yet to be tackled.
 
\begin{itemize}
\item   \emph{Challenge 2} (\emph{High dimensional parameter space | Number of parameters}): The flexibility of the CP is based on providing different covariance parameters for different output levels, therefore, even for a moderate-scale problem, the number of parameters in the covariance function can easily reach hundreds or even thousands. This will lead to significant difficulties in solving the optimization problem to find the maximum likelihood estimator (MLE) for the parameters, specifically under non-convex and highly nonlinear settings.

\item \emph{Challenge 3} (\emph{Negative transfer of knowledge}): The integrative analysis of multiple outputs implicitly assumes that these outputs share some commonalities. However, if this does not hold, negative transfer of knowledge may occur, which leads to decreased performance relative to learning tasks separately~\citep{pan2010survey}. This is specifically important in the CP approach which, unlike separable approaches, implicitly implies that functions have unique features. Even though negative transfer is a very important issue, no research has handled this issue in the context of MGP models.
\end{itemize}

In the current literature, nonseparable modeling using MGCP is prohibitive even for a moderate number of outputs, due to \emph{challenges 1} and \emph{2}. Also, no literature has addressed the negative transfer  of knowledge (\emph{challenge 3}) in MGCP that results from the integrative analysis of outputs that share no commonalities. This article aims to simultaneously overcome these challenges through proposing a regularized pairwise modeling approach for MGCP models. The proposed method has excellent scalability when the number of outputs is large and minimizes the negative transfer of knowledge between uncorrelated outputs. Our approach is based on breaking down the high dimensional MGCP into a group of bivariate GP models, where the shared latent function parameters that facilitate the sharing of information in each bivariate model are penalized to prevent information sharing between outputs with no commonalities. Consistency in estimation and variable selection are then established. Empirical evidence demonstrates that the proposed method can: (1) achieve similar prediction performance as the full multivariate approach when the output dimension is low, (2) outperform the full multivariate approach, with only a fraction of its computational needs, when the output dimension is high, (3) outperform the full multivariate approach when some functions are uncorrelated even when the output dimension is low.  

The rest of the article is organized as follows. Section 2 provides some preliminaries related to the CP construction. In Section 3, we motivate the proposed method through expanding on the proposed challenges. Section 4 introduces our regularized pairwise modeling approach and proves some corresponding statistical properties. The advantageous features of our proposed method are then demonstrated through benchmarking our method with other reference methods in Section 5. Finally, Section 6 concludes this article with discussions. Technical details are deferred to the appendix.

\section{Preliminaries}

Consider a set of $N$ output functions $\bm y (\bm x) = [y_1(\bm x),y_2(\bm x),...,y_N(\bm x)]^t$  from inputs $\bm{x} $ lying in some input space $\mathcal{X} \subset \mathcal{R}^D $. The MGP is defined as 
\begin{equation}
  \label{eq:raed1}
  \begin{gathered}
  \mathcal{Y} (\bm x)=
\begin{bmatrix}
y_1(\bm{x})\\y_2(\bm{x})\\ \vdots \\y_N (\bm{x})
 \end{bmatrix}
 =
 \begin{bmatrix}
f_1 (\bm{x})\\f_2(\bm{x})\\ \vdots \\f_N (\bm{x})
 \end{bmatrix}
 +
 \begin{bmatrix}
\epsilon_1 (\bm{x})\\ \epsilon_2(\bm{x})\\ \vdots \\ \epsilon_N (\bm{x})
 \end{bmatrix}
 =\mathcal{F}(\bm{x}) + \mathcal{E}(\bm{x}),
   \end{gathered}
\end{equation}

\noindent
where the function $\mathcal{F}:\mathcal{R}^D \rightarrow \mathcal{R}^N $ is zero mean multivariate process with covariance $\mbox{cov}^f_{ij}(\bm x,{\bm x}^{\prime})$ $=\mbox{cov}^f_{ij} \big ( f_i(\bm x),f_j({\bm x}^{\prime}) \big)$ for all $\bm x,{\bm x}^{\prime} \in \mathcal{X} $, $i,j \in \mathcal{I}=\{1,2,...,N\}$ and $\epsilon_i(\bm x) \sim \mathcal{N}(0,\sigma^2_i)$ represents additive noise. For the $i$th output the observed data is denoted as $\mathcal{D}_i=\{(\bm{y}_i,\bm{X}_i)\}$, where $\bm y_i = [y_i^1,y_i^2...,y_i^{p_i}]^t$, $y_i^c:=y_i(\bm{x}_{ic})$, $\bm{X}_i=[\bm{x}_{i1},...,\bm{x}_{ip_i}]^t$ and $p_i$ represents the number of observations for output $i$. Also, denote $\bm{X}=[\bm{X}_i^t,...,\bm{X}_N^t]^t$ to be the matrix of all input points. We do not restrict $\bm{X}_i=\bm{X}_j$.  The formulation in (\ref{eq:raed1}) is a general decomposition of an MGP and will reduce to a univariate GP if $N=1$.

\begin{comment} 
includegraphics[width=0.4\textwidth]
\begin{figure*}[!b]
\end{figure*}
\end{comment}

\begin{figure}[H]
\centering
\includegraphics[keepaspectratio=true,scale=0.59]{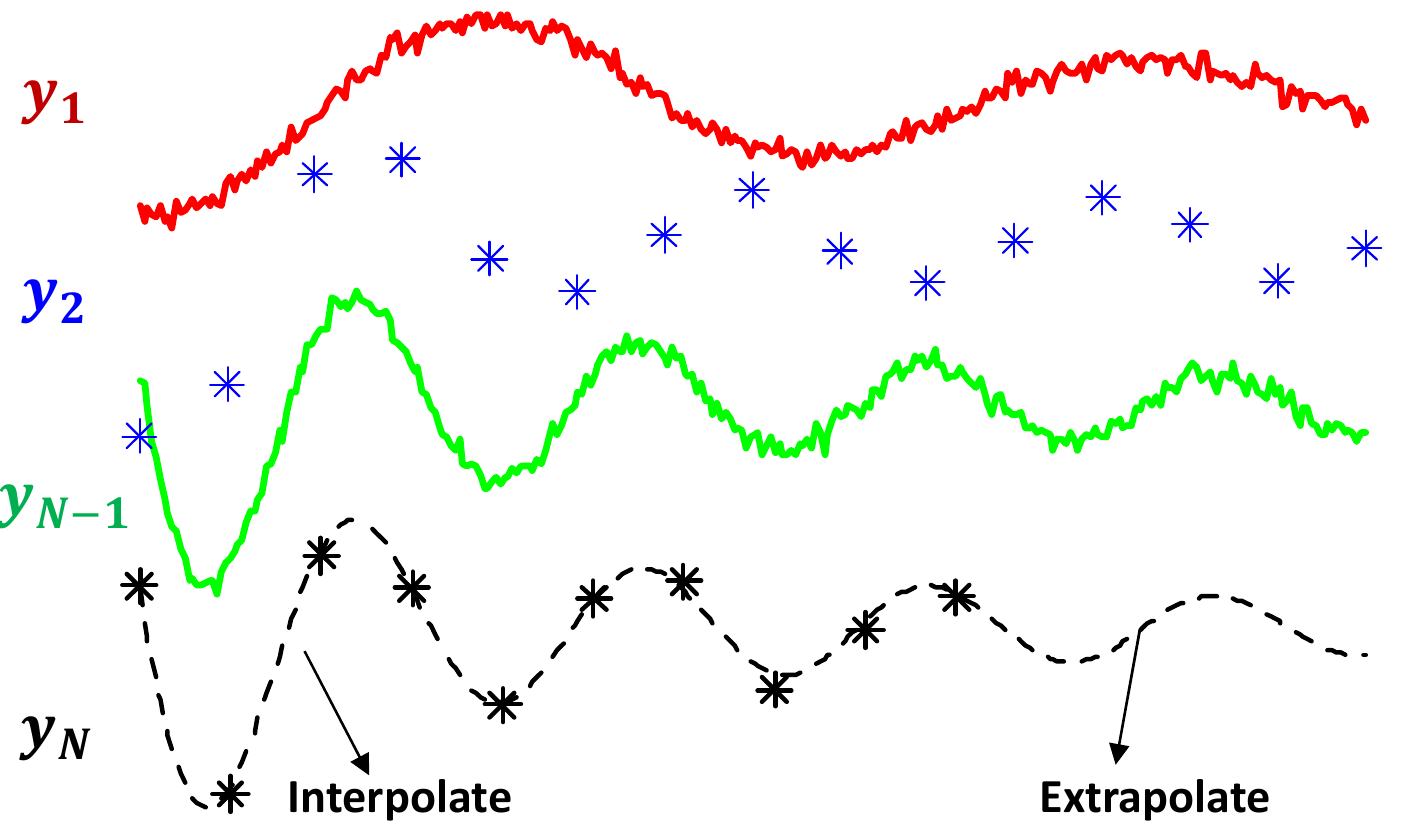}
  \caption{Illustration of General MGP Setting}
  \label{fig:pic1}
\end{figure}

As shown in Figure \ref{fig:pic1}, the underlying principle of the MGP is to borrow strength from a sample of curves, through the shared representation in (\ref{eq:raed1}) which enables integrative analysis, in order to predict individual outputs. For instance, assuming $P=\sum {p_i}$, at any new input $\bm{x}_0 \in \mathcal{X}$ belonging to output $i \in \mathcal{I}$ the joint distribution of the observed values from all outputs and the target function value at the test location $y_i^0=y_i(\bm{x}_0)$ is given by

\begin{equation}
  \label{eq:raed2}
  \begin{gathered}
\left(
\begin{array}{l}
 \bm{y}\\
 y_i^0
\end{array}
\middle\vert
\;\bm{X}
\right)
 \sim \mathcal{N}
 \begin{pmatrix}
\bm{0}_{P+1}, &
\begin{bmatrix}
\bm{C}_{\bm f,\bm f} + \bm \Sigma & \bm{C}_{\bm f,f_i^0}\\[1.5pt]
\bm{C}^t_{\bm f,f_i^0} & C_{f_i^0,f_i^0} + \sigma^2_i 
\end{bmatrix}
 \end{pmatrix},
   \end{gathered}
\end{equation}

\noindent
where $\bm{y}=[\bm{y}_1^t,\bm{y}_2^t,...,\bm{y}_N^t]^t$ are the noisy observed targets corresponding to the latent function values $\bm{f}=[\bm{f}_1^t,\bm{f}_2^t,...,\bm{f} _N^t]^t$, $f_i^c:=f_i(\bm{x}_{ic})$ such that $\bm{f}_i=f_i(\bm{X}_i)$, $\bm{C}_{\bm f,\bm f} \in \mathcal{R}^{P\times P}$ is the covariance matrix relating all input points for all outputs with $\mbox{cov}^f_{ij}(\bm x,{\bm x}^{\prime})$, $\bm \Sigma=diag[\sigma^2_1 \bm{I}_{p_1},...,\sigma^2_N\bm{I}_{p_N} ]$ is a block diagonal matrix in which the $i$th block corresponds to a $p_i \times p_i$ matrix, $\bm{C}_{\bm f,f_i^0}=[\bm{C}^t_{\bm{f}_1,f_i^0},...,\bm{C}^t_{\bm{f}_N,f_i^0}]^t$; $\bm{C}_{\bm{f}_c,f_i^0}=[\mbox{cov}^f_{ic}(\bm{x}_0,\bm{x}_{c1}),...,\mbox{cov}^f_{ic}(\bm{x}_0,\bm{x}_{cp_c})]^t$ and $C_{f_i^0,f_i^0}=\mbox{cov}^f_{ii}(\bm{x}_0,{\bm{x}_0})$ where $f_i^0:=f_i(\bm{x}_0)$. Following multivariate normal theory, the predictive distribution of $y_i(\bm{x}_0)$ denoted as $pr(\cdot|\bm y)$  is given as
\begin{equation}
  \label{eq:raed3}
pr(y_i(\bm{x}_0)|\bm y)=\mathcal{N} \Big(    \bm{C}^t_{\bm f,f_i^0} (\bm{C}_{\bm f,\bm f} + \bm \Sigma)^{-1}\bm{y},\ C_{f_i^0,f_i^0}+ \sigma^2_i-\bm{C}^t_{\bm f,f_i^0} (\bm{C}_{\bm f,\bm f} + \bm \Sigma)^{-1}\bm{C}_{\bm f,f_i^0} \Big).
\end{equation}

The mean in (\ref{eq:raed3}) is the empirical best linear unbiased estimator (EBLUP) of $y_i(\bm{x}_0)$, while the variance is divided into three parts, the first is the variance of the variable under study, $\sigma^2_i$ represents additive noise and the last part is the variance reduction due to the EBLUP approximation \citep{stein1991universal}.

As shown from (\ref{eq:raed3}), the sharing of information is achieved through  $\mbox{cov}^f_{ij}(\bm x,{\bm x}^{\prime})$ which models the variations both within and across different outputs, to capture their relatedness and improve prediction accuracy. This covariance function is typically assumed to belong to a known parametric class \citep{calder2007some} of covariance functions $\{\mbox{cov}^f_{ij}(\cdot,\cdot \ ; \bm{\theta}_f),\bm{\theta}_f \in \Theta_f  \}$, where $\forall \bm{\theta}_f \in \Theta_f, \ \mbox{cov}^f_{ij}(\cdot,\cdot \ ; \bm{\theta}_f)$ is a valid positive semidefinite covariance function, such that $\Theta_f$ is a set that contains the true parameters $\bm{\theta}^{\ast}_f$ for the covariance of $\bm{f}(\bm{x})$. Although in univariate settings there are many well known valid autocovariance functions, however, in the MGP it is extremely challenging to define cross-covariance functions that result in valid covariance matrices \citep{boyle2007gaussian}. 

An alternative to  directly parametrizing a covariance function is to construct a GP, $f_i:\mathcal{R}^D \rightarrow \mathcal{R}$, through convolving a Gaussian white noise process $X(\bm{x})$ with a smoothing kernel $K_i(\bm{x})=\alpha_i k_i(\bm{x})$ where $\alpha_i \in \mathcal{R}$ and $k_i:\mathcal{R}^D \rightarrow \mathcal{R}$. Since the base process is a GP, and a convolution is a linear operator on a function, then the convolved process is also a GP \citep{thiebaux1987spatial,barry1996blackbox,higdon2002space}.

\begin{equation}
\label{eq:raed4}
f_i(\bm{x})=K_i(\bm{x}) \star X(\bm{x})=\int_{-\infty}^{\infty}K_i(\bm{x} - \bm{u}) X(\bm{u}) d \bm{u} \ ,
\end{equation}

\noindent
where $\star$ denotes a convolution, $\mbox{cov}(X_i(\bm{u}),X_i(\bm{u}^{\prime}))=\delta(\bm{u}-\bm{u}^{\prime} )=\delta(\bm{u}^{\prime}-\bm{u})=\delta_{\bm{u}\bm{u}^{\prime}}$ and $\delta$ is the Dirac delta function. This approach is equivalent to applying a stable linear filter, where the output $f_i(\bm{x})$ is a weighted integral over the input signal $X(\bm{x})$, weighted according to the impulse response $K_i(\bm{x})$. This requires the filter to be stable, where the output is bounded for all bounded input signals \citep{haykin2008communication}, i.e. for a positive real finite number $a$, $|X(\bm{x})|\leq a \implies  |f_i(\bm{x})| \leq a \int_{-\infty}^{\infty}|k_i(\bm{u})| d \bm{u}$. Therefore the only restriction for constructing a valid GP is that the impulse response/kernel is absolutely integrable  $\int_{-\infty}^{\infty}|k_i(\bm{u})| d \bm{u} < \infty$. Some applications and extensions of the CP approach for the single ouput case are presented in \cite{wikle2002kernel} and \cite{calder2007some}.

Under the CP construction as shown in (\ref{eq:raed4}), if we share the same latent function $X(\bm{x})$, across multiple outputs $f_i(\bm{x}), i\in \mathcal{I}$, then all outputs can be expressed as a jointly distributed GP, i.e. MGP \citep{higdon2002space}. The resulting covariance function will then only depend on the displacement vector $\bm{d}=\bm{x}-{\bm x}^{\prime} \in \mathcal{R}^D$ and is given as 

\begin{equation}
\label{eq:raed5}
\mbox{cov}^f_{ij}(\bm x,{\bm x}^{\prime})=\int_{-\infty}^{\infty}K_i(\bm{x} - \bm{u}) K_j({\bm x}^{\prime} - \bm{u}) d \bm{u} = \int_{-\infty}^{\infty}K_i(\bm{u}) K_j(\bm{u}-\bm{d}) d \bm{u}.
\end{equation}

As shown in (\ref{eq:raed5}), instead of directly parametrizing a positive semi-definite covariance function we only need to specify the parameters of a smoothing kernel, where the resulting covariance function must be valid by construction. Therefore, the key advantages is that we can exploit influence of multiple latent functions, $X_q(\bm{x})$ where $q\in\{1,2,..,Q\}$, to share information across different output levels through different covariance parameters encoded in the kernels $K_{qi}(\bm{x})$ as shown in (\ref{eq:raed6})~\citep{fricker2013multivariate,alvarez2012kernels}.

\begin{equation}
\label{eq:raed6}
f_i(\bm{x})=\sum_{q=1}^{Q} K_{qi}(\bm{x}) \star X_q(\bm{x})=\sum_{q=1}^{Q}\int_{-\infty}^{\infty}K_{qi}(\bm{x} - \bm{u}) X_q(\bm{u}) d \bm{u}.
\end{equation}

\section{Challenges and Motivation}

In this section, we expand on the MGCP challenges and present the motivation for our proposed method.

\subsection{Computational complexity}

As shown in the previous section, the MGCP is fully parametrized through the kernel parameters $\bm{\theta}_f$ and measurement noise $\bm{\sigma}=\{\sigma_1,...,\sigma_N\}$. Now denote $\bm{\theta}=\{\bm{\theta}^t_f,\bm{\sigma}^t\}^t$ and all the observed data as $\mathcal{D}=\{\mathcal{D}_i,...,\mathcal{D}_N\}$, then the likelihood of the joint MGCP is given as

\begin{equation}
\begin{aligned}
\mathcal{L}(\bm{\theta};\mathcal{D})=(2\pi)^{-P/2} |\bm{C}_{\bm f,\bm f} + \bm \Sigma|^{-1/2}\times \mbox{exp}(-\bm{y} (\bm{C}_{\bm f,\bm f} + \bm \Sigma)^{-1} \bm{y}^t/2).
\end{aligned}
\label{eq:raed7}
\end{equation}

As shown in (\ref{eq:raed7}), learning from the likelihood requires the inversion of $\bm{C}_{\bm f,\bm f} + \bm \Sigma$ and calculating its determinant at each step/iteration. The covariance matrix of the CP, assuming $p$ observations for each of the $N$ outputs, scales as $Np$ leading to $\mathcal{O}(P^3)=\mathcal{O}(N^3p^3)$ computational complexity and $\mathcal{O}(N^2p^2)$ storage. Therefore, CP modeling approaches are plagued by extremely high computational loads and numerical issues. In addition, more input points are usually required for multivariate outputs, resulting in a further increase in the complexity. Although separable approaches consider multiple outputs however their restrictive covariance functions often lead to structured covariances which are easily manipulated \citep{qian2008gaussian,zhou2011simple}. On the other hand, some recent approaches have tried to tackle this computational issue in the context of nonseparable covariances~\citep{alvarez2011computationally,nguyen2014collaborative}. For instance, \cite{alvarez2011computationally} proposed an inducing variable approximation similar to the well-known partially independent training conditional (PITC) approximation \citep{quinonero2005unifying}. Their key assumption is that that outputs $\{y_i(\bm{x}):i \in \mathcal{I}\}$ would be conditionally independent if all latent functions $X_q(\bm{u})$ are observed at some inducing locations, i.e. $pr(\{y_i(\bm{x})\}_{i=1}^N|\{X_q(\bm{u})\}_{q=1}^Q)= \prod_{i=1}^N pr(y_i(\bm{x})|\{X_q(\bm{u})\}_{q=1}^Q)$. This assumption leads to the inversion of a block diagonal covariance matrix which reduces the complexity to $\mathcal{O}(Np^3)$ which matches that of modeling with $N$ independent GPs. Further, \cite{kontar2017nonparametricb,kontar2017nonparametrica} assumed training output as independent and proposed sharing latent functions $X_q(\bm{u})$ only between test and training outputs. This lead to a block arrowhead covariance matrix which also reduced the complexity of learning from the likelihood to that of modeling with independent GPs. It crucial to note that despite the reduced computational complexity due to the manipulation of the covariance matrix, we are still required to estimate an extremely large number of parameters  even with a moderate number of outputs. This challenge is further discussed in the next subsection.

\subsection{Number of parameters}

Despite the disadvantages associated with high computational complexity, the main drawback, which renders nonseparable modeling prohibitive even for a moderate $N$, is the large number of the parameters that need to be estimated. This large number of parameters is a direct consequence of the CP construction which inherits its flexibility from the ability to share information through different kernels resulting in different covariance parameters for different output levels. It is clear from (\ref{eq:raed3}) that prediction accuracy is greatly dependent on the parameter estimates which are obtained from minimizing the negative log-likelihood function $\ell(\bm{\theta};\mathcal{D})=-\mbox{log} \ \mathcal{L}(\bm{\theta};\mathcal{D})$. Up to an additive constant, $\ell(\bm{\theta};\mathcal{D})= \frac{1}{2} \langle \bm{Y},(\bm{C}_{\bm f,\bm f} + \bm \Sigma)^{-1}\rangle + \frac{1}{2}\mbox{log}|\bm{C}_{\bm f,\bm f} + \bm \Sigma| $ where $\langle \bm{A} , \bm{A}^{\prime} \rangle=trace(\bm{A} \bm{A}^{\prime})$ and $\bm{Y}=\bm{y}\bm{y}^t$. Since, $\ell(\bm{\theta};\mathcal{D})$ is highly nonlinear and non-convex in  $\bm{\theta}$, its minimization in a high dimensional parameter space is extremely challenging, time consuming and suboptimal as one should anticipate poor parameter estimates in such high dimensions \citep{rasmussen2004gaussian,tajbakhsh2014sparse,mardia1989multimodality}.

\begin{comment}
and suboptimal as standard optimization technique can easily get trapped in local minima
\end{comment}

\begin{figure}[H]
\centering
\includegraphics[keepaspectratio=true,scale=0.8]{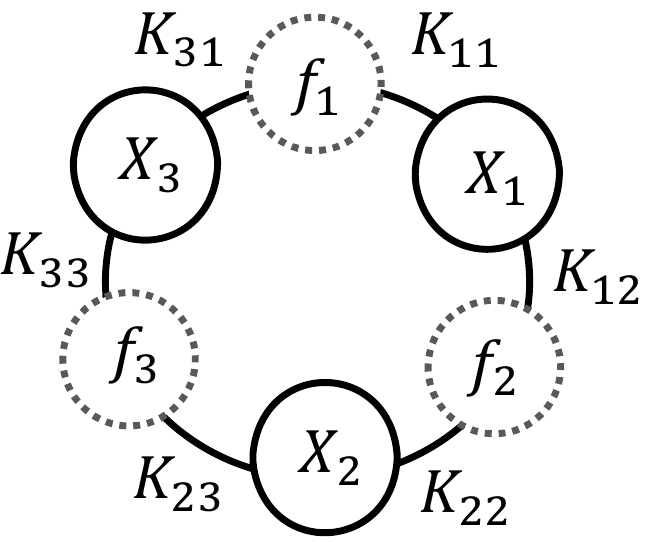}
  \caption{CP  model structure example}
  \label{fig:pic2}
\end{figure}

Furthermore, as shown in (\ref{eq:raed6}), the number of parameters depends on $Q$ which is the number of latent functions induced in the model. For instance, consider the case in Figure \ref{fig:pic2} where we share latent functions between each pair of outputs, i.e. $Q=N(N-1)/2$ resulting in $N(N-1)$ kernels. This is comparable to two way interaction effects in ANOVA where~\citep{wonnacott1990introductory} only pairwise interactions are considered through the shared latent GP. Now assume the kernels $K_{iq}(\bm{x})=\alpha_{qi} k_{qi}(\bm{x})$ follow the most commonly used exponential kernel, $k_{qi}(\bm{x})=\mbox{exp}((\bm{x}-{\bm x}^{\prime})^t \bm{\Lambda}_{qi} (\bm{x}-{\bm x}^{\prime}))$ where $\bm{\Lambda}_{qi}$ is a $D \times D$ positive definite diagonal matrix allowing different length scales for each dimension. As a result the total number of parameters in the model is $N(N-1)(1+D)  + N$ where the first part $N(N-1)$ represents the number of kernels multiplied by the number of parameters $(1+D)$ in each kernel, while the second part $N$ represents the number of parameters in $\bm{\sigma}$. Note that this case is a bit conservative as we use the exponential kernel which is able to provide a large degree of flexibility with a small number of hyper parameters \citep{rasmussen2004gaussian,colosimo2014profile}. Even for a moderate scale case where $N=30$ and $D=1$, we are required to estimate 1770 parameters under a non-convex setting. In another case, considering the more restrictive approach in \cite{li2016pairwise} and \cite{li2018pairwise} where $Q=1$ and all outputs possess the same noise, i.e. $\sigma=\sigma_1=,..,=\sigma_N$, the  number of parameters still scales as $N(N+2D+1)/2+1$, therefore for $N=30$ and $D=1$ we are estimating 991 parameters. In conclusion, obtaining good estimates in such a high dimensional parameters space is an impractical task, for this reason the practical applications of the MGCP are limited. Note that once the parameters are learned, prediction complexity at any new test point $\bm{x}_0$ is $\mathcal{O}(Np)$ for mean and $\mathcal{O}(N^2p^2)$ for variance, which can be done rather efficiently. 

\subsection{Negative transfer}
\begin{figure}[H]
\centering
\makebox[\textwidth]{\includegraphics[keepaspectratio=true,width=\textwidth]{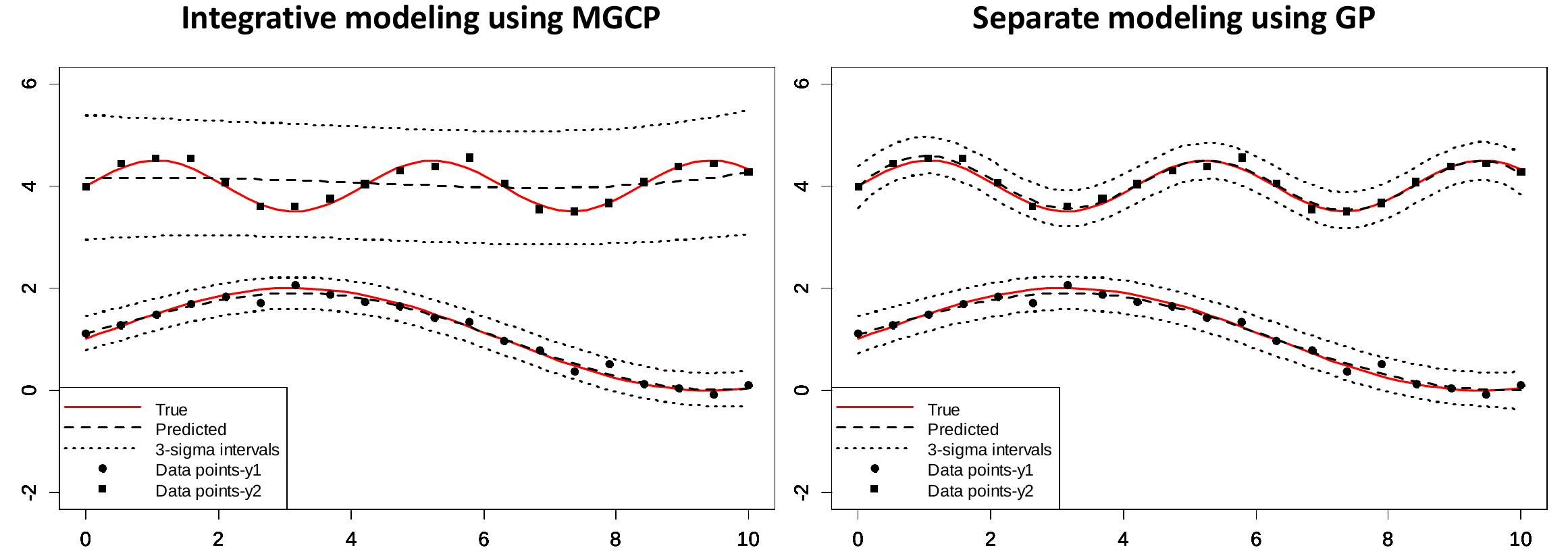}}
  \caption{Illustration of negative transfer in MGCP }
  \label{fig:pic3}
\end{figure}

As previously mentioned, a major challenge in the MGP which to the best of our knowledge has not been tackled yet, is the negative transfer of knowledge that occurs when we integratively analyze outputs that share no commonalities. Similar to the second challenge, negative transfer is specifically important in nonseperabale approaches which are used when functions have unique features. We start with an example to illustrative the impact of negative transfer in MGCP models. Consider a simple case with two outputs and a one dimensional input $x \in \mathcal{X} \subset \mathcal{R}$. The outputs are generated according to $y_1(x)=1+\mbox{sin}(x)+\epsilon_1(x)$ and $y_2(x)=4+0.5\mbox{sin}(1.5x)+\epsilon_2(x)$ and $x \in [0,10]$. The number of observations per signal is $p_1=p_2=20$ evenly spaced points and the measurement noise is set as $\sigma_1=\sigma_2=0.1$. We analyze this data separately using a univariate GP applied to each output and integratively using an MGCP. In the univariate GP, we assume a Gaussian/squared exponential covariance where $\mbox{cov}^f_{ii} (x,x^{\prime})=\alpha_i^2\mbox{exp}(-d^2/(4\nu^2))$. In the MGCP we adopt the covariance function in \cite{fricker2013multivariate} and \cite{li2018pairwise}, where $\mbox{cov}^f_{ij} (x,x^{\prime})=\alpha_i\alpha_j\sqrt{2|\nu_i\nu_j| / (\nu_i^2+\nu_j^2}) \mbox{exp}(-d^2/(2\nu_i^2+2\nu_j^2))$. Results are shown in Figure \ref{fig:pic3}.

The results clearly indicate that separate modeling of each function is significantly better than there integrative analysis. This is specifically clear for $y_2$, which interestingly, was predicted using a larger length scale than the truth due to sharing of information with a smoother function $y_1$. Note that this problem occurred in our dense input example ($p_i=20$) which implies that the challenge of negative transfer becomes exceedingly significant with sparse data. For these reasons, and since negative transfer is a crucial issue in MGP models, the main goal of this article is to handle the negative transfer of knowledge while maintaining the scalability of the model. 

\subsection{Motivation: Pairwise estimation and the precision matrix}

In this article we propose a pairwise distributed estimation scheme motivated by both the distributed estimation literature for univariate GP models \citep{tresp2000Bayesian,deisenroth2015distributed} and pairwise modeling of longitudinal profiles \citep{fieuws2006pairwise,li2016pairwise}. The proposed approach is based on distributing MGCP estimation through bivariate GP submodels which are individually estimated. Predictions are then made through combining predictions from the bivariate models within a Bayesian framework. Not only does this approach scale to arbitrarily large datasets by parallelization, also each bivariate model can be efficiently built with a limited of parameters and a small-scale covariance matrix. 

While, pairwise modeling seems to address computational challenges, negative transfer remains an important issue. Interestingly, \emph{pairwise modeling turns out to possess unique characteristics which  allows us to tackle the challenge of negative transfer}. While few literature~\citep{williams2009multi} have aimed to establish the number of latent functions to be shared, such approaches do not imply avoiding negative transfer. As we will show in this subsection, information sharing and independent predictions can only be avoided through sparsity on the precision matrix. However, since MGCP models are based on modeling the covariance through latent functions not the precision matrix then we can only control the precision matrix under specific structures. It is clear from (\ref{eq:raed3}) that prediction accuracy for the GP/MGP is dependent on the inverse covariance matrix, also known as precision matrix $\bm{\Omega}=(\bm{C}_{\bm f,\bm f} + \bm \Sigma)^{-1} \in \mathcal{R}^{P \times P}$. The precision matrix carries conditional independence information. This matrix consists of block matrices $\bm{\Omega}_{ij}\in \mathcal{R}^{p_i \times p_j}$, where the $(c,c^{\prime})^{th}$ entry of each block is denoted as $\Omega_{ij}^{c,c^{\prime}}$. One can directly show that $\mbox{cov}(y_i^c,y_j^{c^{\prime}}|\ddot{\bm{y}}) = 0 $ if and only if $\Omega_{ij}^{c,c^{\prime}}=0$, where $\ddot{\bm{y}}=\{\bm{y}\}/ \{y_i^c,y_j^{c^{\prime}}\}$ ($\bm{y}$ excluding $y_i^c$ and $y_j^{c^{\prime}}$). Thus, conditionally independent variables lead to zero entries in the precision matrix \citep{whittaker2009graphical}. As mentioned previously, GP/MGP models are characterized through a positive semidefinite covariance function (ex: $\mbox{cov}^f_{ij}(\bm x,{\bm x}^{\prime})$) rather than a conditional covariance function to generate the precision matrix. However, the remarks below illustrate some useful properties in the case of a bivariate GP. 

\noindent\\
\textbf{Lemma 1.} (\emph{Multivariate}) \emph{ Given that $pr(\bm{y}|\bm{X},\bm{\theta})=\mathcal{N}(\bm{0}_P,\bm{\Omega}^{-1})$, then $\bm{\Omega}_{ij}=\bm{0}$  if and only if the multivariate Gaussian random vectors $\bm{y}_i$ and $\bm{y}_j$ are conditionally independent, i.e. $\mbox{cov}(y_i^c,y_j^{c^{\prime}}|\ddot{\bm{y}})=0$ for every $c \in \{1,..,p_i\}$ and $c^{\prime} \in \{1,..,p_j\}$}.\\

\noindent
\textbf{Lemma 2.} (\emph{Bivariate}) \emph{ Given that $pr(\bm{y}_i,\bm{y}_j|\bm{X}_1,\bm{X}_2,\bm{\theta}^{\prime})= \mathcal{N} \begin{pmatrix} \bm{0}_{p_i+p_j}, &\begin{pmatrix}
\bm{\Omega}_{ii} & \bm{\Omega}_{ij}\\
\bm{\Omega}_{ij}^t & \bm{\Omega}_{jj} 
\end{pmatrix}^{-1} \end{pmatrix}$, then  $\bm{\Omega}_{ij}=\bm{0}$ if and only if, the multivariate Gaussian random vectors $\bm{y}_i$ and $\bm{y}_j$ are independent, i.e. $\mbox{cov}(y_i^c,y_j^{c^{\prime}})=0$ for every $c \in \{1,..,p_i\}$ and $c^{\prime} \in \{1,..,p_j\}$}.\\

\noindent
A brief proof of both Lemmas and further references are provided in Appendix A. The key conclusion from the Lemmas is that \emph{through pairwise modeling, we are able to control the precision matrix through parameters in the covariance function used to construct the bivariate GP}. 

The utilization of pairwise modeling, distributed estimation and this direct mapping from the covariance function to the precision matrix is detailed in the following sections.

\section{Model Development}

The proposed framework presents a flexible alternative that can scale to a large number of outputs while avoiding the negative transfer of knowledge. The nature of our proposed pairwise approach circumvents any need to find or establish latent functions between pairs. While within each pair, model selection is automatically done through our regularization approach which is consistently able to infer whether information should be shared or not. In Section 4.1, we establish our pairwise model based on a CP construction. Our pairwise scheme is based on distributing the estimation of the high dimensional MGCP into bivariate GP's which are individually built. An MGCP with $N$ outputs as a result decomposes into $N(N-1)/2$ pairwise submodels to predict all outputs. However, for the sake of notational simplicity, and building on Figure \ref{fig:pic1}, we focus on predicting one output through sharing information from the remaining $N-1$ outputs as shown in Figure \ref{fig:pic4} below. In Section, 4.2 we provide some statistical guarantees for the proposed method. Section 4.3, provides a direct approach to applying our regularized pairwise approach to separable modeling. Finally, Section 4.4, provides the methodology to combine predictions from the bivariate submodels.

\begin{figure}[H]
\centering
\includegraphics[keepaspectratio=true,scale=0.6]{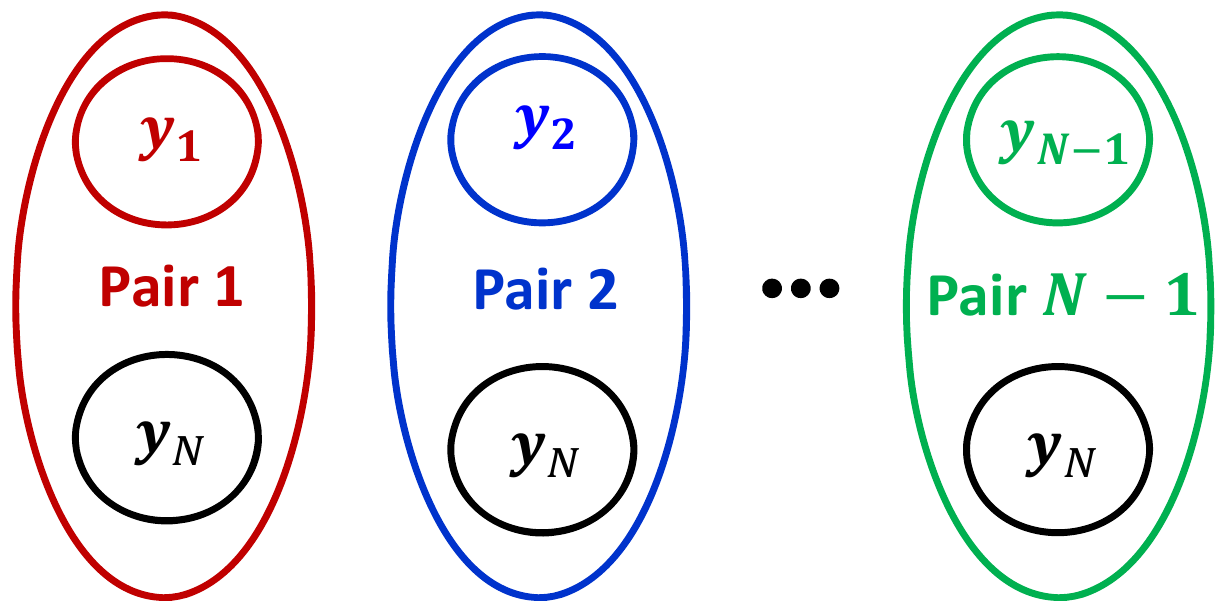}
  \caption{Paired submodels}
  \label{fig:pic4}
\end{figure}

\subsection{The pairwise and regularized MGCP}

\begin{figure}[H]
\centering
\makebox[\textwidth]{\includegraphics[keepaspectratio=true,width=\textwidth]{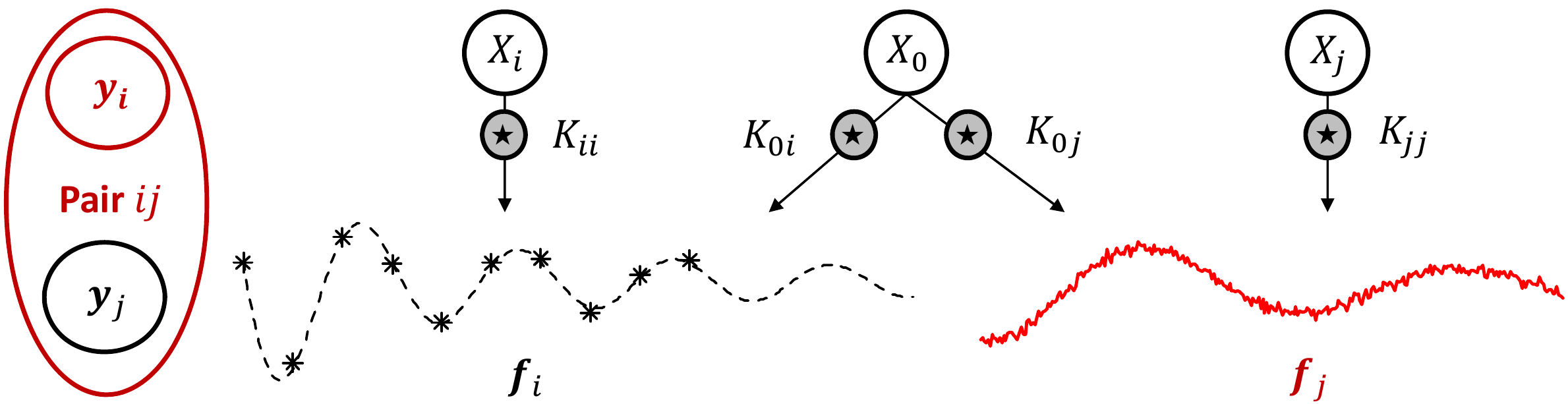}}
  \caption{Bivariate submodel structure}
  \label{fig:pic5}
\end{figure}

Based on (\ref{eq:raed1}) for each pairwise submodel we have
$
\begin{bmatrix}
y_i(\bm{x})\\y_j(\bm{x})\\
 \end{bmatrix}
 =
 \begin{bmatrix}
f_i(\bm{x})\\f_j(\bm{x})\\
 \end{bmatrix}
 +
 \begin{bmatrix}
\epsilon_i(\bm{x})\\ \epsilon_j(\bm{x})\\ 
 \end{bmatrix}
$, for $i,j \in \mathcal{I}$, where the input data is $\mathcal{D}_i=\{(\bm{y}_i,\bm{X}_j)\}$ and $\mathcal{D}_j=\{(\bm{y}_j,\bm{X}_j)\}$. Further, we assume that $\bm{y}_{ij}=[\bm{y}_i^t,\bm{y}_j^t]^t$ represents the noisy observations corresponding to the latent function values $\bm{f}_{ij}=[\bm{f}_i^t,\bm{f}_j^t]^t$. In order to capture both the unique properties of each output and their interdependence we construct both $f_i(\bm{x})$ and $f_j(\bm{x})$ as the sum of a latent function unique to each output and a shared latent function which facilitates the sharing of information. The model structure is shown in Figure \ref{fig:pic5}.

As shown in Figure \ref{fig:pic5}, independent features are encoded through dependence on a latent function that has no effect on the other output $(X_i,X_j)$ while dependent features are encoded through the common dependence on $X_0$.

Following the CP construction in (\ref{eq:raed6}), we have that $y_i(\bm{x})=f_i(\bm{x})+\epsilon_i(\bm{x})=K_{ii}(\bm{x}) \star X_i(\bm{x})+ K_{0i}(\bm{x}) \star X_0(\bm{x})+\epsilon_i(\bm{x})$, similarly, $y_j(\bm{x})=K_{jj}(\bm{x}) \star X_j(\bm{x})+ K_{0j}(\bm{x}) \star X_0(\bm{x})+\epsilon_j(\bm{x})$. This model is quite flexible, as it provides both shared and unique latent functions for both outputs. Based on this modeling framework, the cross covariance function between the two outputs is given as 

\begin{equation}
\label{eq:raed8}
\mbox{cov}^y_{ij}(\bm x,{\bm x}^{\prime})=\mbox{cov}^f_{ij}(\bm x,{\bm x}^{\prime})+\mbox{cov}^{\epsilon}_{ij}(\bm x,{\bm x}^{\prime})=\mbox{cov}^f_{ij}(\bm x,{\bm x}^{\prime})+\sigma_i^2 \tau_{ij}\tau_{x,x^{\prime}} ,
\end{equation}

\noindent
where, $\tau_{ij}$ is the Kronecker delta function, which is equal to one if $i=j$, and is zero otherwise. In a more general case to that of Section 2, we define the latent functions $\{X_q:q=0,i,j\}$ as $\mbox{cov}(X_q(\bm{u}),X_q(\bm{u}^{\prime}))=\xi_q^2 \delta_{\bm{u}\bm{u}^{\prime}}$ where $\xi_q \in \mathcal{R}$. Then given the fact that $X_q(\bm{u})$ and $X_{q^{\prime}}(\bm{u}^{\prime})$ are independent latent functions which only covary if $q=q^{\prime}$ and $\bm{u}=\bm{u}^{\prime}$ and utilizing the commutativity of the convolution and the "sifting" property of the Dirac delta function, i.e. $\int f(\bm{u})\delta(\bm{u}-\bm{x})d\bm{z}=f(\bm{x})$, we have that
\begin{equation}
\begin{split}
\mbox{cov}^f_{ij}(\bm x,{\bm x}^{\prime})& = E(f_i(\bm{x})f_j(\bm{x}^{\prime}))
\\&=E\Big(\sum_{q=\{0,i\}} \int_{-\infty}^{\infty}K_{qi}(\bm{x} - \bm{u}) X_q(\bm{u}) d \bm{u} \times \sum_{q^{\prime}=\{0,j\}} \int_{-\infty}^{\infty}K_{qj}(\bm{x}^{\prime} - \bm{u}^{\prime}) X_{q^{\prime}}(\bm{u}^{\prime}) d \bm{u}^{\prime} \Big)
\\&=\sum_{q=\{0,i,j\}} \int_{-\infty}^{\infty}\int_{-\infty}^{\infty} K_{qi}(\bm{u})K_{qj}(\bm{u}^{\prime}) E(X_q(\bm{x} - \bm{u})X_q(\bm{x}^{\prime} - \bm{u}^{\prime}) d\bm{u} d\bm{u}^{\prime}
\\&=\sum_{q=\{0,i,j\}} \xi_q^2 \int_{-\infty}^{\infty}\int_{-\infty}^{\infty} K_{qi}(\bm{u})K_{qj}(\bm{u}^{\prime}) \delta(\bm{x}-\bm{u}-\bm{x}^{\prime} + \bm{u}^{\prime})  d\bm{u} d\bm{u}^{\prime}
\\&=\sum_{q=\{0,i,j\}} \xi_q^2 \int_{-\infty}^{\infty} K_{qi}(\bm{u})K_{qj}(\bm{u}-\bm{d}) d\bm{u}=\sum_{q=\{0,i,j\}} \xi_q^2 \int_{-\infty}^{\infty} K_{qi}(\bm{u}+\bm{d})K_{qj}(\bm{u}) d\bm{u} .
\end{split}
\label{eq:raed9}
\end{equation}

Note that the derivation in (\ref{eq:raed9}) is a wrapper function for $\mbox{cov}^f_{ii}(\bm x,{\bm x}^{\prime})$, $\mbox{cov}^f_{jj}(\bm x,{\bm x}^{\prime})$ and $\mbox{cov}^f_{ij}(\bm x,{\bm x}^{\prime})$. Based on our model framework illustrated in Figure \ref{fig:pic5} and due to the shared dependence on only $X_0$, the autocovariance and cross covariance functions simplify as
 
\begin{equation}
\begin{cases} 
\mbox{cov}^f_{ii}(\bm x,{\bm x}^{\prime})=\sum_{q=\{0,i\}} \xi_q^2 \int_{-\infty}^{\infty} K_{qi}(\bm{u})K_{qi}(\bm{u}-\bm{d}) d\bm{u}
\\
\mbox{cov}^f_{jj}(\bm x,{\bm x}^{\prime})=\sum_{q=\{0,j\}} \xi_q^2 \int_{-\infty}^{\infty} K_{qj}(\bm{u})K_{qj}(\bm{u}-\bm{d}) d\bm{u}
\\
\mbox{cov}^f_{ij}(\bm x,{\bm x}^{\prime})=\xi_0^2 \int_{-\infty}^{\infty} K_{0i}(\bm{u})K_{0j}(\bm{u}-\bm{d}) d\bm{u}
\end{cases} 
\label{eq:raed10}
\end{equation}

This modeling framework is generic in terms of the choice of the kernel function. However, a general purpose kernel  can be constructed through assuming the kernels follow a Gaussian form. As mentioned previously, the Gaussian kernel is the most common choice of kernels due to its flexibility and correspondence to Bayesian linear regression with an infinite number of basis functions \citep{rasmussen2004gaussian}. Further, similar constructions using such kernels have been utilized in both a GP and MGP setting \citep{paciorek2004nonstationary,alvarez2011computationally}.

Now assume the kernels $K_{qi}(\bm{x})=\alpha_{qi} (4\pi)^\frac{D}{4}|\bm{\Lambda}_{qi}|^{-\frac{1}{4}} \mathcal{N}(\bm{x}|\bm{0},\bm{\Lambda}_{qi}^{-1} )$ to be scaled Gaussian kernels. Also, denote $\mathcal{N}(\bm{x}|\bm{\mu}_{qi},\bm{\Lambda}_{qi}^{-1} )\mathcal{N}(\bm{x}|\bm{\mu}_{qj},\bm{\Lambda}_{qj}^{-1} )=\mathcal{N}(\bm{\mu}_{qi}-\bm{\mu}_{qj}|\bm{0},\bm{\Lambda}_{qi}^{-1}+\bm{\Lambda}_{qj}^{-1}) \mathcal{N}(\bm{x}|\tilde{\bm{\mu}},\tilde{\bm{\Lambda}} )$ where $\tilde{\bm{\Lambda}}^{-1}=(\bm{\Lambda}_{qi}+\bm{\Lambda}_{qj})^{-1}$ and $\tilde{\bm{\mu}}=\tilde{\bm{\Lambda}}^{-1}(\bm{\Lambda}_{qi}\bm{\mu}_{qi}+\bm{\Lambda}_{qj}\bm{\mu}_{qj})$, to be the identity for the product of two Gaussian distributions. Then, we have that
\begin{equation}
\begin{split}
\mbox{cov}^f_{ij}(\bm x,{\bm x}^{\prime})& =\sum_{q=\{0,i,j\}} \xi_q^2 \omega_{ij}^q  \int_{-\infty}^{\infty} \mathcal{N}(\bm{u}|-\bm{d},\bm{\Lambda}_{qi}^{-1} ) \mathcal{N}(\bm{u}|\bm{0},\bm{\Lambda}_{qj}^{-1} ) d\bm{u}
\\&=\sum_{q=\{0,i,j\}} \xi_q^2 \omega_{ij}^q  \int_{-\infty}^{\infty} \mathcal{N}(\bm{u}|-\bm{d},\bm{\Lambda}_{qi}^{-1}+\bm{\Lambda}_{qj}^{-1} ) \mathcal{N}(\bm{u}|\tilde{\bm{\mu}},\tilde{\bm{\Lambda}} ) d\bm{u}
\\&=\sum_{q=\{0,i,j\}} \xi_q^2 \omega_{ij}^q  \mathcal{N}(\bm{d}|\bm{0},\bm{\Lambda}_{qi}^{-1}+\bm{\Lambda}_{qj}^{-1} )
=\sum_{q=\{0,i,j\}} \xi_q^2 \tilde{\omega}^q_{ij}  \mbox{exp}( -\frac{1}{2} \bm{d}^t {\bm{\Phi}^{q}_{ij}}^{-1} \bm{d} ),
\end{split}
\label{eq:raed11}
\end{equation}

\noindent
where $\omega_{ij}^q=\alpha_{qi} \alpha_{qj}(4\pi)^\frac{D}{2}|\bm{\Lambda}_{qi}|^{-\frac{1}{4}}|\bm{\Lambda}_{qj}|^{-\frac{1}{4}}$, $\tilde{\omega}^q_{ij}=2^\frac{D}{2}\alpha_{qi} \alpha_{qj} |\bm{\Lambda}_{qi}|^{\frac{1}{4}}|\bm{\Lambda}_{qj}|^{\frac{1}{4}}/|\bm{\Lambda}_{qi}+\bm{\Lambda}_{qj}|^\frac{1}{2}$, and ${\bm{\Phi}^{q}_{ij}}^{-1}=(\bm{\Lambda}_{qi}^{-1}+\bm{\Lambda}_{qj}^{-1})^{-1}=\bm{\Lambda}_{qi}(\bm{\Lambda}_{qi}+\bm{\Lambda}_{qj})^{-1}\bm{\Lambda}_{qj}$. A nice feature of (\ref{eq:raed11}), is that the marginal process  i.e. $\mbox{cov}^f_{ii}(\bm x,{\bm x}^{\prime})=\sum_{q=\{0,i\}} \xi_q^2 \alpha_{qi}^2 \mbox{exp}( -\frac{1}{4} \bm{d}^t \bm{\Lambda}_{qi} \bm{d} )$ has the most common Gaussian covariance function resulting from the convolution of two Gaussian kernels. Therefore, (\ref{eq:raed11}) can be viewed as the extension of the Gaussian covariance function to the multivariate case. Once again we note that (\ref{eq:raed11}) is a wrapper function where for $i \neq j$, we have that $\mbox{cov}^f_{ij}(\bm x,{\bm x}^{\prime})=\xi_0^2 \tilde{\omega}^0_{ij}  \mbox{exp}( -\frac{1}{2} \bm{d}^t {\bm{\Phi}^{0}_{ij}}^{-1} \bm{d} )$.

Now, we let $\bm{\theta}_{f_{ij}} \in \Theta_{f_{ij}}$ represent the parameters in $\mbox{cov}^f_{ij}(\bm x,{\bm x}^{\prime})$ where $\Theta_{f_{ij}}$ is a set that contains the true parameters $\bm{\theta}^{\ast}_{f_{ij}}$, and we denote $\bm{\theta}_{ij}^t=\{\bm{\theta}_{f_{ij}}^t,\bm{\sigma}_{ij}^t\}^t$, where $\bm{\sigma}_{ij}=\{\sigma_i,\sigma_j\}^t$,  to be the set of all parameters in the bivariate submodel. Then, given our CP formulation and following (\ref{eq:raed8}), the marginal density of the bivariate submodel is expressed as $pr(\bm{y}_i,\bm{y}_j|\bm{X}_1,\bm{X}_2,\bm{\theta}_{ij})=$
\begin{equation}
\label{eq:raed12}
\int pr(\bm{y}_{ij}|\bm{f}_{ij})pr(\bm{f}_{ij}|\bm{\theta}_{f_{ij}})d\bm{f}_{ij}=\int \prod_{c=1}^{p_i} pr(y_i^c|f_i^c) \prod_{c^{\prime}=1}^{p_j} pr(y_j^{c^{\prime}}|f_j^{c^{\prime}})pr(\bm{f}_{ij}|\bm{\theta}_{f_{ij}})d\bm{f}_{ij}, 
\end{equation}

\noindent
where $pr(\bm{f}_{ij}|\bm{\theta}_{f_{ij}})=\mathcal{N}(\bm{0}_{P^{\prime}},\bm{C}_{{\bm f}_{ij},{\bm f}_{ij}})$, $pr(\bm{y}_{ij}|\bm{f}_{ij})=\mathcal{N}(\bm{0}_{P^{\prime}},\bm{\Sigma}_{ij})$ and $\mathrm{p}=p_i+p_j$. Therefore $pr(\bm{y}_i,\bm{y}_j|\allowbreak\bm{X}_1,\bm{X}_2,\bm{\theta}_{ij})=\mathcal{N}(\bm{0}_{\mathrm{p}},\bm{C}_{{\bm f}_{ij},{\bm f}_{ij}}+\bm{\Sigma}_{ij})$ where

\begin{equation}
\label{eq:raed13}
\bm{C}_{{\bm f}_{ij},{\bm f}_{ij}}+ \bm{\Sigma}_{ij}= 
\begin{pmatrix} 
\bm{C}_{\bm{f}_i,\bm{f}_i} & \bm{C}_{\bm{f}_i,\bm{f}_j}\\
\bm{C}_{\bm{f}_i,\bm{f}_j}^t & \bm{C}_{\bm{f}_j,\bm{f}_j} 
 \end{pmatrix}+
 \begin{pmatrix} 
\sigma^2_i \bm{I}_{p_i} & \bm{0}\\
\bm{0} & \sigma^2_j \bm{I}_{p_j} 
 \end{pmatrix} \ ,
\end{equation}

\noindent
such that $\bm{C}_{{\bm f}_{ij},{\bm f}_{ij}} \in \mathcal{R}^{\mathrm{p}\times \mathrm{p}} $ is the covariance matrix relating all input points of outputs $i$ and $j$ with $\mbox{cov}^f_{ij}(\bm x,{\bm x}^{\prime})$ in (\ref{eq:raed10}) and (\ref{eq:raed11}). As previously mentioned, parameter estimates are obtained from minimizing the negative log-likelihood function $\ell(\bm{\theta}_{ij};\mathcal{D}_i,\mathcal{D}_j)=-\mbox{log} \  pr(\bm{y}_i,\bm{y}_j|\bm{X}_1,\bm{X}_2,\bm{\theta}_{ij})$. Denoting $\bm{Y}_{ij}=\bm{y}_{ij}\bm{y}^t_{ij}$,  up to an additive constant, the bivariate likelihood and its derivatives are given as

\begin{equation}
\label{eq:raed14}
\ell(\bm{\theta}_{ij};\mathcal{D}_i,\mathcal{D}_j)= \frac{1}{2} \langle \bm{Y}_{ij},(\bm{C}_{{\bm f}_{ij},{\bm f}_{ij}}+ \bm{\Sigma}_{ij})^{-1}\rangle + \frac{1}{2}\mbox{log}|\bm{C}_{{\bm f}_{ij},{\bm f}_{ij}}+ \bm{\Sigma}_{ij}|  \ .
\end{equation}

Further, through denoting $\bm{C}_{ij}=\bm{C}_{{\bm f}_{ij},{\bm f}_{ij}}+ \bm{\Sigma}_{ij}$, $\bm{\Psi}_{ij}=\bm{C}_{ij}^{-1}\bm{y}_{ij}$ and $\bm{\Xi}_{nm}=\frac{\partial  \bm{C}_{ij} }{\partial \theta_{ij}^{(n)}}  \bm{C}_{ij}^{-1}  \frac{\partial \bm{C}_{ij}}{\partial \theta_{ij}^{(m)}}$ the gradient and second derivatives with respect to any parameter $\theta_{ij}^{(n)} \in \bm{\theta}_{ij}$ are then given as (in Appendix B we expand on ${\partial  \bm{C}_{ij} }/{\partial \theta_{ij}^{(n)}}$ )

\begin{gather*}
\frac{\partial \ell}{\partial \theta_{ij}^{(n)}}=\frac{1}{2} \bigg\langle \bm{\Psi}_{ij} \bm{\Psi}_{ij}^t - \bm{C}_{ij}^{-1}, \frac{\partial \bm{C}_{ij}}{\partial \theta_{ij}^{(n)}}   \bigg\rangle \\
\frac{\partial^2 \ell}{\partial \theta_{ij}^{(n)} \partial \theta_{ij}^{(m)} } =\frac{1}{2} \bigg\langle \bm{\Psi}_{ij} \bm{\Psi}_{ij}^t - \bm{C}_{ij}^{-1}, \bigg( \frac{\partial^2 \bm{C}_{ij}}{\partial \theta_{ij}^{(n)} \partial \theta_{ij}^{(m)}} -   \bm{\Xi}_{nm} \bigg)  
\bigg\rangle -  
\frac{1}{2} \bigg\langle \bm{\Psi}_{ij} \bm{\Psi}_{ij}^t, \bm{\Xi}_{mn}  \bigg\rangle \ .
\end{gather*}

The computational complexity of learning from the bivariate likelihood in (\ref{eq:raed14}) is reduced to $\mathcal{O}((2p)^3)$. More importantly, the total number of parameters in the model is reduced to $4(1+D) + 2 + 3$ where $4$ represents the number of kernels ($K_{ii},K_{jj},K_{0i},K_{0j}$) multiplied by the number of parameters $(1+D)$ in each kernel, while 2 represents $\sigma_i$ and $\sigma_j$, and 3 represents the parameters ($\xi_0,\xi_i,\xi_j$) of the latent functions ($X_0,X_i,X_j$) respectively. Note that reductions in parameter number can be done through assuming $K_{0i}=K_{0j}$ or $\sigma_i=\sigma_j$. As shown distributed estimation using bivariate submodels, is able to handle both the computational complexity and large parameter number to be estimated. Also, all pairwise models can be parallelized and thus our model can scale to an arbitrarily large number of outputs by parallelization.

Now, recall Figure (\ref{fig:pic5}), $X_0$, defined by $\xi_0$, represents the latent function which facilitates the sharing of information between outputs $i$ and $j$. Therefore, in order to handle negative transfer of knowledge we use the bivariate likelihood in (\ref{eq:raed14}) but with $\xi_0$ penalized. Following Lemma 2, we will show in the following section that shrinking $\xi_0$ decreases the cross correlation amongst the outputs and that $\xi_0=0$ ensures that each output is predicted independently. The penalized negative log-likelihood function is defined as 

\begin{equation}
\label{eq:raed15}
\ell_\mathbb{P}(\bm{\theta}_{ij};\mathcal{D}_i,\mathcal{D}_j,\lambda)=\ell(\bm{\theta}_{ij};\mathcal{D}_i,\mathcal{D}_j)+\mathbb{P}_\lambda(|\xi_0|) \ ,
\end{equation}

\noindent
where $\mathbb{P}_\lambda(\xi_0)$ is a penalty function. Different types of penalty functions can be used, examples include: ridge penalty $\mathbb{P}_\lambda(|\xi_0|)=\lambda \xi_0^2$, $\ell_1$ penalty $\mathbb{P}_\lambda(|\xi_0|)=\lambda |\xi_0|$, bridge penalty $\mathbb{P}_\lambda(|\xi_0|)=\lambda |\xi_0|^{0 <  \cdot < 1}$, and scad penalty  $\mathbb{P}_\lambda(|\xi_0|)=\lambda |\xi_0|$ if $|\xi_0| \leq \lambda$, $(\xi_0^2-2\gamma\lambda|\xi_0|+\lambda^2)/(2\gamma-2)$ if $\lambda<|\xi_0| \leq \gamma\lambda$, $\lambda^2(\gamma+1)/2$ if $|\xi_0| > \gamma\lambda$. The tuning parameter $\lambda$ ($\lambda$ and $\gamma$ in Scad) has an important effect on predictions. For instance in the Lasso, as $\lambda$ increases, $\xi_0$ shrinks to zero. Typically, the optimal tuning parameter is found using a grid search such as generalized cross validation (GCV), specifically $b$-fold GCV \citep{friedman2001elements}.  The GCV is based on splitting the data into $b$ groups. For a given $\lambda$, first, we exclude one group as the testing dataset, and the rest $b-1$ groups are used as the training data, from which predictions at test locations are obtained. This is repeated $b$ times for each group. The tuning parameter is then chosen as the minimizer of some accuracy measure (ex: absolute error, mean squared error, etc..) based on the GCV procedure.

\subsection{Statistical properties}
We now discuss some of the structural properties of our model. We first introduce the main theorem which guarantees that our regularized model in (\ref{eq:raed15}) is able to avoid the negative transfer of knowledge.\\

\noindent
\textbf{Theorem 1.} \emph{Suppose that $\xi_0=0$, then the predictive distribution of the bivariate model at any new input $\bm{x}_0 \in \mathcal{X}$ reduces to that of a univariate model where $pr(y_i(\bm{x}_0)|\bm{y}_{ij})=pr(y_i(\bm{x}_0)|\bm{y}_{i})$ and $pr(y_j(\bm{x}_0)|\bm{y}_{ij})=pr(y_j(\bm{x}_0)|\bm{y}_{j})$ such that
$$
\begin{cases} 
pr(y_i(\bm{x}_0)|\bm{y}_{ij})=\mathcal{N} \Big(    \bm{C}^t_{\bm{f}_i,f_i^0} \bm{\Omega}_{ii} \bm{y}_i,\ C_{f_i^0,f_i^0}+ \sigma^2_i-\bm{C}^t_{\bm{f}_i,f_i^0} \bm{\Omega}_{ii} \bm{C}_{\bm{f}_i,f_i^0} \Big) \\
pr(y_j(\bm{x}_0)|\bm{y}_{ij})=\mathcal{N} \Big(    \bm{C}^t_{\bm{f}_j,f_j^0} \bm{\Omega}_{jj}\bm{y}_j,\ C_{f_j^0,f_j^0}+ \sigma^2_j-\bm{C}^t_{\bm{f}_j,f_j^0} \bm{\Omega}_{jj} \bm{C}_{\bm{f}_j,f_j^0} \Big)
\end{cases} 
$$
where for $c\in\{i,j\}$, $\bm{\Omega}_{cc}=(\bm{C}_{\bm{f}_c,\bm{f}_c} + \sigma^2_c \bm{I}_{p_c})^{-1}\in\mathcal{R}^{p_c \times p_c}$, $\bm{C}_{\bm{f}_c,f_c^0}=[\mbox{cov}^f_{cc}(\bm{x}_0,\bm{x}_{c1}),...,\allowbreak \mbox{cov}^f_{cc}(\bm{x}_0,\bm{x}_{cp_c})]^t$,  $C_{f_c^0,f_c^0}=\mbox{cov}^f_{cc}(\bm{x}_0,{\bm{x}_0})$ and $\mbox{cov}^f_{cc}(\bm x,{\bm x}^{\prime})= \xi_c^2 \int_{-\infty}^{\infty} K_{cc}(\bm{u})K_{cc}(\bm{u}-\bm{d}) d\bm{u}$}. \\

The proof is detailed in Appendix C. The key feature of this theorem is that penalizing only one variable, in our initial parameter set $\xi_0\in \bm{\theta}_{f_{ij}} \subset \bm{\theta}_{ij}$,  will lead to separating the bivariate model into two models equivalent to the univariate GP established through a CP in (\ref{eq:raed4}). Our regularization approach is flexible to any specified kernel function and not based on the Gaussian covariance derived in (\ref{eq:raed11}), where theorem 1 holds for any valid kernel function $K_{iq}$. We note that the result of theorem 1, is based on the fact that for $c\in\{i,j\}$ as $\xi_0 \rightarrow 0$
$$\begin{cases} 
\mbox{cov}^f_{ij}(\bm x,{\bm x}^{\prime})=\xi_0^2 \int_{-\infty}^{\infty} K_{0i}(\bm{u})K_{0j}(\bm{u}-\bm{d}) d\bm{u} \rightarrow 0 \\
\mbox{cov}^f_{cc}(\bm x,{\bm x}^{\prime})=\sum_{q=\{0,c\}} \xi_q^2 \int_{-\infty}^{\infty} K_{qi}(\bm{u})K_{qi}(\bm{u}-\bm{d}) d\bm{u} \rightarrow \xi_c^2 \int_{-\infty}^{\infty} K_{cc}(\bm{u})K_{cc}(\bm{u}-\bm{d}) d\bm{u} .
\end{cases}$$

This is important to note since non-sparse penalties such as the ridge penalty can still minimize the negative transfer of knowledge through shrinking $\xi_0$. This however comes at the expense of variable selection implied in sparse penalties.

Next, we discuss some asymptotic properties of our regularized bivariate model. In order to investigate the asymptotic properties of our regularized model, we first need to examine the properties of $\bm{\theta}_{ij}$ obtained from minimizing the unpenalized likelihood in (\ref{eq:raed14}). Here we note that one  advantage of our model is that for both the penalized  $\ell_{\mathbb{P}}(\bm{\theta}_{ij})$ and unpenalized likelihood $\ell(\bm{\theta}_{ij})$  we are minimizing over the same set of parameters $\bm{\theta}_{ij}$ since $\xi_0 \in \bm{\theta}_{ij}$. Based on mild regularity conditions for dependent observations, it has been shown that the maximum likelihood estimator of $\bm{\theta}_{ij}$ is $r_{\mathrm{p}}$ consistent. Refer to Appendix D, for more details \citep{basawa1980statistical,basawa1976asymptotic,shi2011gaussian}. Now for the penalized model, let $\bm{\theta}^{\ast \ t}_{ij}=\{\bm{\theta}^{\ast \ t}_{f_{ij}},\bm{\sigma}^{\ast \ t}_{ij}\}^t$ be the true parameter values corresponding to  $\bm{\theta}_{ij}^t=\{\bm{\theta}_{f_{ij}}^t,\bm{\sigma}_{ij}^t\}^t$, and let $\hat{\bm{\theta}}_{ij}$ be the estimated parameters obtained from minimizing $\ell_\mathbb{P}(\bm{\theta}_{ij})$. Hence $\xi_0^{\ast}$ and $\hat{\xi}_0$ respectively represent the true and estimated value of $\xi_0$ . For the penalty function $\mathbb{P}_\lambda(|\xi_0|)$, we assume that the penalty is non-negative; $\mathbb{P}_\lambda(|\xi_0|)\geq 0 $ and $\mathbb{P}_\lambda(0)= 0$, and that larger coefficients are penalized no less than smaller ones; $\mathbb{P}_\lambda(|\xi_0^{\prime}|) \geq \mathbb{P}_\lambda(|\xi_0|)$ if $|\xi_0^{\prime}| \geq |\xi_0|$. These are typical assumptions and are satisfied by the aforementioned penalties \citep{fan2001variable}. Further, we assume that the first and second derivatives of $\mathbb{P}_\lambda(|\xi_0|)$ are continuous at $\xi_0^{\ast} \neq 0$.  We next provide two theorems that establish parameter estimation and selection consistency. The theroems provide similar results as in \cite{fan2001variable}, but defined within our model specifications and based on dependent observations. Note that the \raisebox{0.3ex}{``$\prime$"} notation on a function implies a derivative. \\

\noindent
\textbf{Theorem 2.} \emph{If $z_2=\mbox{max}\{|\mathbb{P}^{\prime\prime}_\lambda(|\xi_0^{\ast}|)|:\xi_0^{\ast} \neq 0\} \rightarrow 0$, then there exits a local minimzer $\hat{\bm{\theta}}_{ij}$ for $\ell_\mathbb{P}(\bm{\theta}_{ij})$, such that $||\hat{\bm{\theta}}_{ij}-\bm{\theta}^{\ast}_{ij}||=O(r^{-1}_{\mathrm{p}}+z_1)$, where $z_1=\mbox{max}\{\mathbb{P}^{\prime}_\lambda(|\xi_0^{\ast}|):\xi_0^{\ast} \neq 0\}$   }. \\

\noindent
\textbf{Theorem 3.} \emph{Assume that $\xi_0^{\ast}=0$ and the parameters $\ddot{\bm{\theta}}_{ij}=\{\bm{\theta}_{ij}\}/ \{\xi_0\}$ satisfy $r_{\mathrm{p}}$ consistency in theorem 2. Then if $\underset{\mathrm{p} \rightarrow \infty}{lim \ inf} \ \underset{\xi_0 \rightarrow 0^{+}}{lim \ inf} \ \frac{1}{\lambda} \mathbb{P}^{\prime}_\lambda(\xi_0) > 0$, $\lambda \rightarrow 0$ and ${\mathrm{p}\lambda}/{r_{\mathrm{p}}} \ \rightarrow \infty$ as $\mathrm{p} \rightarrow \infty$, we have that $\underset{\mathrm{p} \rightarrow \infty}{lim} pr(\hat{\xi}_0=0)=1$}.\\ 

The proof for theorems 2 and 3 is detailed in Appendix D. In the above theorem  ``\emph{lim \ inf}" denoted the infimum of the limit points. It is clear from theorem 2 and 3, that if we choose a proper tuning parameter $\lambda$ and penalty function $\mathbb{P}_\lambda(|\xi_0|)$ there exists an $r_{\mathrm{p}}$ consistent estimator for the penalized likelihood  $\ell_\mathbb{P}(\bm{\theta}_{ij})$, which possesses the sparsity property  $\hat{\xi}_0=0$, i.e. asymptotically performs as well as knowing that ${\xi}_0=0$ beforehand. This result is also known as an oracle property which provides consistency in variable selection \citep{donoho1995adapting}. Here variable selection implies selecting whether functions should be predicted independently or not, a stated in theorem 1. 

\subsection{Application to separable covariance}
In this section, we provide a direct approach to applying our regularized pairwise approach to separable modeling. Fortunately, since separable models provide a simplified alternative to nonseparable models, avoiding negative transfer can be readily accomplished. As previously mentioned, the covariance function in a separable model is of the form $\mbox{cov}^f_{ij}(\bm x,{\bm x}^{\prime})=T_{ij}\mbox{cov}(\bm x,{\bm x}^{\prime})$, where $T_{ij}$ is the between-output covariance matrix and $\mbox{cov}(.,.)$ is a covariance function over inputs $\bm{x}\in \mathcal{R}^D$, the same for all outputs. For $\mbox{cov}^f_{ij}(\bm x,{\bm x}^{\prime})$ to be a valid covariance function, it is required that $T_{ij}=\{t_{c,c^{\prime}}\}$ be a positive definite matrix with unit diagonal elements (PDUDE) \citep{zhou2011simple}. Therefore, in a bivariate case with two outputs $i$ and $j$, $T_{ij}=\begin{pmatrix} 1 & t_{ij}\\ t_{ij} & 1 \end{pmatrix}$, such that $t_{ij}$ measures the correlation between output $i$ and $j$. It is interesting to note that the covariance between outputs only varies through $t_{ij}$, this is why outputs in separable functions are instantaneously mixed as they are directly derived by a scaling or a rotation to an output space of dimension $D$. Inspired by Theorem 1 and Appendix C, one can directly show that $t_{ij}=0$ ensures that each output is predicted independently. Therefore, in separable modeling, we only need to adjust the penalty function to $\mathbb{P}_\lambda(|t_{ij}|)$, and optimize the penalized likelihood while restricting $T_{ij}$ to be PDUDE. It is interesting to mention here the intrinsic coregionalization (IC) model ~\citep{helterbrand1994universal} which is a simplified version of the LMC previously mentioned in the introduction. The IC is a separable construction that reduces to independent predictions over each output under an isotopic data case and if outputs are modeled as noise free. Unfortunately, despite its ability to avoid negative transfer, such a model cannot make use of commonalities across outputs. 

\subsection{Combining the predictions}
Without loss of generality,  we focus on predicting output $N$ through sharing information from the remaining $N-1$ outputs as shown in Figure \ref{fig:pic4}. Based on (\ref{eq:raed3}), for each submodel in Figure \ref{fig:pic4}, the predictive equation for any new input $\bm{x}_0 \in \mathcal{X}$ for output $N$ is expressed as\begin{equation}
  \label{eq:raed16}
pr(y_N(\bm{x}_0)|\bm{y}_{iN})=\mathcal{N} \Big( \bm{C}^t_{\bm{f}_{iN},f_N^0} \bm{\Omega}_{iN} \bm{y}_{iN},\ C_{f_i^N,f_i^N}+ \sigma^2_N-\bm{C}^t_{\bm{f}_{iN},f_N^0} \bm{\Omega}_{iN} \bm{C}_{\bm{f}_{iN},f_N^0} \Big),
\end{equation}
where $i \in \mathcal{I}^{-N}= \{1,...,N-1\}$, $\bm{\Omega}_{iN}=(\bm{C}_{{\bm f}_{iN},{\bm f}_{iN}}+ \bm{\Sigma}_{iN})^{-1}$, $\bm{C}_{\bm{f}_{iN},f_N^0}=[\bm{C}^t_{\bm{f}_i,f_N^0},\bm{C}^t_{\bm{f}_N,f_N^0}]^t$ and $\bm{C}_{\bm{f}_c,f_N^0}=[\mbox{cov}^f_{Nc}(\bm{x}_0,\bm{x}_{c1})\allowbreak,...,\mbox{cov}^f_{Nc}(\bm{x}_0,\bm{x}_{cp_c})]^t$ for  $c\in\{i,N\}$. Our goal is to efficiently combine the predictions from the $N-1$ bivariate submodels to form an overall result. To this end we utilize the product of GP experts (PoE) model, used in univariate GP's, however implemented within the specifications of our pairwise model \citep{ng2014hierarchical,deisenroth2015distributed}.  Here, we aim to combine predictions from $N-1$ ``experts", where each expert is a regularized bivariate GP. The PoE model combines the predictions by the product of all expert predictions. In our pairwise model, the PoE implies that $\bar{pr}(y_N(\bm{x}_0)|\bm{y})=\prod_{c=1}^{N-1} pr(y_N(\bm{x}_0)|\bm{y}_{cN})$. PoE models are straightforward and theoretically appealing as each expert is weighted by the inverse covariance, therefor experts which
are uncertain about their predictions are automatically weighted less than experts that are
certain about their predictions. However, a major shortcoming of PoE models is that as $N$ increases, the combined prediction tends to be overconfident. For instance, assume that all functions in $\mathcal{I}^{-N}$ are exactly equivalent, then we have that $pr(y_N(\bm{x}_0)|\bm{y}_{cN})=\mathcal{N}(\mathcal{M},\mathcal{V}) \ \forall c\in \mathcal{I}^{-N} $ for some mean $\mathcal{M}$ and variance $\mathcal{V}$. Therefore, $\bar{pr}(y_N(\bm{x}_0)|\bm{y})=\mathcal{N}(\bar{\mathcal{M}}=\mathcal{M},\bar{\mathcal{V}}=\mathcal{V}/(N-1))$ and as $N \rightarrow \infty \implies \bar{\mathcal{V}} \rightarrow 0 $. Naturally, in such a case we would want $\bar{pr}(y_N(\bm{x}_0)|\bm{y})=\mathcal{N}(\bar{\mathcal{M}}=\mathcal{M},\bar{\mathcal{V}}=\mathcal{V})$. To this end, we weight the contributions of each expert with a weight $\beta_c=1/(N-1)$ for $c \in \mathcal{I}^{-N}$. As a result, given that $pr(y_N(\bm{x}_0)|\bm{y}_{cN})=\mathcal{N}(\mathcal{M}_c,\mathcal{V}_c) \ \forall c\in \mathcal{I}^{-N}$, and following the identity that the product of Gaussian distributions is Gaussian, we have that $\bar{pr}(y_N(\bm{x}_0)|\bm{y})=\mathcal{N}(\bar{\mathcal{M}},\bar{\mathcal{V}})$, where

\begin{equation}
\label{eq:raed17}
\bar{\mathcal{V}}^{-1}=\sum_{c=1}^{N-1} \beta_c \mathcal{V}^{-1}_c, \quad \bar{\mathcal{M}}=\bar{\mathcal{V}}\sum_{c=1}^{N-1} \beta_c \mathcal{V}_c^{-1}\mathcal{M}_c \ ,
\end{equation}

This efficient closed form inference for combining the bivariate models is independent of the computational graph, and, consequently, facilitates the ability to scale to arbitrarily large datasets by parallelization, where each bivariate model is efficiently built with a limited of parameters and a small-scale covariance matrix. Note that (\ref{eq:raed17}) is similar to the log opinion pool model \citep{heskes1998selecting} and the Generalized product of experts \citep{cao2014generalized,deisenroth2015distributed}, therefore, the key feature of the PoE model is still retained as experts that are uncertain about their predictions are  weighted less, also, since $\sum_{c\in \mathcal{I}^{-N}} \beta_c=1$, then it ensures a consistent model that falls back to the prior. Some slight modifications to the traditional PoE have been also proposed such as the Bayesian Committee Machine (BCM) or the robust BCM \citep{tresp2000Bayesian,schwaighofer2003transductive} . The BCM is based on adjusting the variance of the unweighted PoE, by a prior variance $pr(y_N(\bm{x}_0))$, while the robust BCM adjusts the BCM with weights similar to those in our model. However, the BCM is based upon assuming a block diagonal covariance where all experts share the same parameters, which hinders its application in our model.  \\

\section{Numerical Case Studies}

We conduct case studies to demonstrate the advantageous features of our regularized and distributed multivariate Gaussian convolution process denoted as MGCP-RD. In Section 5.1, we discuss benchmarked methods and the general setting for our numerical case studies. Then Section 5.2 uses simulated functions to demonstrate the performance of the proposed method under four different model settings. Further an illustrative example is provided in Section 5.2.

\subsection{General settings}

In this section, we discuss the settings used to assess the MGCP-RD performance using simulated data. To evaluate the performance of our proposed method, we randomly generate $N$ signals from different model settings, in which  the first $N-1$ outputs are used as a training set, while the $N$th output is  selected as the testing function. We repeat the study for $W=1000$ times. For each replication, we use the mean absolute error (MAE) between the true signal value and its predicted value at $p_{\mbox{test}}=50$ points  as the criterion to evaluate our prediction accuracy. We then report the distribution of the MAE across the $W$ simulations using a group of boxplots, with respective means represented as black dots.  The MAE values are denoted as prediction errors in all boxplots. Kernel parameters are obtained through minimizing the negative log-likelihood using a scaled conjugate gradient algorithm (\cite{li2016pairwise,alvarez2011computationally,rasmussen2004gaussian}). For the MGCP-RD, we distribute the computations over only two systems, where each system was responsible for sequentially fitting $(N-1)/2$ of the bivariate models. All computations are done on R-3.2.2 in a 64-bit Windows 7 setting. Further, in our simulation studies, we benchmark our method with four other reference methods for comparison: 1) The individual GP established using a CP, denoted as GCP, where the test function is fitted separately \citep{rasmussen2004gaussian}; 2) The full MGCP model, denoted as MGCP, described in Section 3.2; 3) The inducing variable approximation, denoted as MGCP-I, which tackles the computational complexity challenge \citep{alvarez2011computationally,alvarez2012kernels}; 4) The pairwise model for longitudinal profiles, denoted as MGCP-P \citep{fieuws2006pairwise,li2016pairwise,li2018pairwise}. To provide consistent results we utilize the Gaussian kernel in Section 4.1 for each of the benchmarked methods. In the GCP, this kernel reduces to the Gaussian/squared exponential covariance function in the univariate case. For the MGCP-I and since no specific latent structure is proposed we use that of the full MGCP in Section 3.2. However, for the MGCP-P, we utilize the proposed latent structure in the paper which only involves one common latent function for each pairwise model. Finally, throughput the numerical study we use a 3-fold cross validation method to find the tuning parameters for our approach.

\subsection{Results}
We simulate functions from three different settings to demonstrate the benefits of the MGCP-RD. The model settings and results are shown below.
\subsubsection{Setting I}
\begin{figure}[H]
\centering
\includegraphics[keepaspectratio=true,scale=0.6]{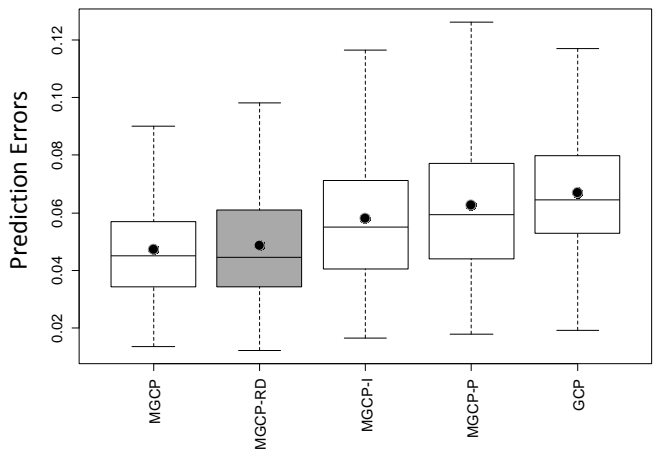}
  \caption{Setting I results}
  \label{fig:pic6}
\end{figure}

In this setting, we aim to compare the performance of the MGCP-RD in a simple case with few number of outputs ($N=5$) and no negative transfer. In order to establish a setting with no negative transfer, the multivariate output model for the $N$ curves are generated according to the same functional form $y_i(x)=1+\mbox{sin}(x)+\epsilon_i(x)$ for $x \in [0,10]$ and $i \in \mathcal{I}$. The number of observations per signal is $p=p_1=,..,=p_N=10$ evenly spaced points, the test points are evenly spaced across $[0,10]$ and measurement noise standard deviation is set to $\sigma=\sigma_1=,..,=\sigma_N=0.1$ for all outputs. For the MGCP-I, all $p$ design points are used as  inducing variables. Also, in this setting we implement the ridge penalization. The importance of this case is that the three challenges of nonseparable modeling can be readily handled in a low dimensional setting with no negative transfer. Therefore, this scenario is able to evaluate the performance of the MGCP-RD relative to the full MGCP. The results are shown in Figure \ref{fig:pic6}.

The results in Figure (\ref{fig:pic6}) indicate that there is an insignificant difference in the prediction error between the full model (MGCP) and our proposed method. First, amongst the models that considered multiple outputs the MGCP-P had the worst performance. The MGCP-P is based on averaging parameter estimates from paired submodels. Such an approach is extremely dangerous due to the multimodality of the GP/MGP likelihood \citep{mardia1989multimodality}. As mentioned in \cite{rasmussen2004gaussian}, different parameters estimates correspond to different interpretations of the data, however, predictions will be moderately effected as the Bayesian correspondence in the GP implies that predictions should pass through or close to the design points. As will be shown in later settings, this approach of averaging parameter estimates becomes specifically dangerous with high noise levels, large parameter space and outputs with varying forms and characteristics  where parameter estimates fluctuate widely between different submodels and iterations. Second, and following the intuition why we average predictions rather than parameter estimates, Figure (\ref{fig:pic7}) below shows the advantageous features of the weighted PoE model. In Figure (\ref{fig:pic7}), we compare the MGCP-RD results with MGCP-RD prediction errors before averaging the $N-1$ pairwise submodels from each iteration (denoted as BIVARIATE). Note that since we consider the marginal errors from each submodel in the BIVARIATE then we have $W(N-1)$ errors compared to MGCP-RD with G errors. As shown in the figure, the straightforward  mechanism of PoE models, where experts that are uncertain about their predictions are automatically weighted less by the inverse covariance, provides both a simple yet efficient solution for distributed modeling.

\begin{figure}[H]
\centering
\includegraphics[keepaspectratio=true,scale=0.4]{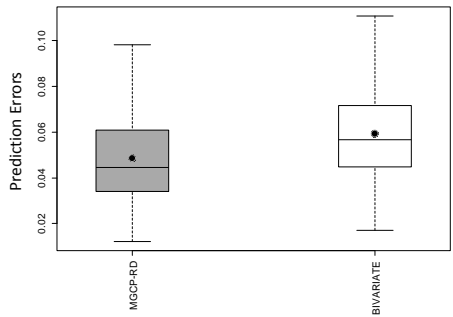}
  \caption{PoE results}
  \label{fig:pic7}
\end{figure}

\subsubsection{Setting II}
\begin{figure}[H]
\centering
\makebox[\textwidth]{\includegraphics[keepaspectratio=true,width=\textwidth]{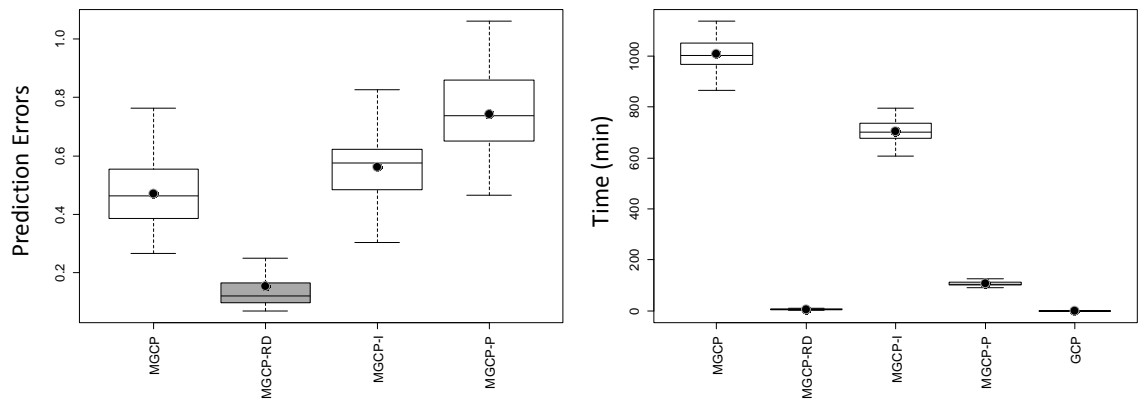}}
  \caption{Setting II results}
  \label{fig:pic8}
\end{figure}

\begin{figure}[H]
\centering
\includegraphics[keepaspectratio=true,scale=0.35]{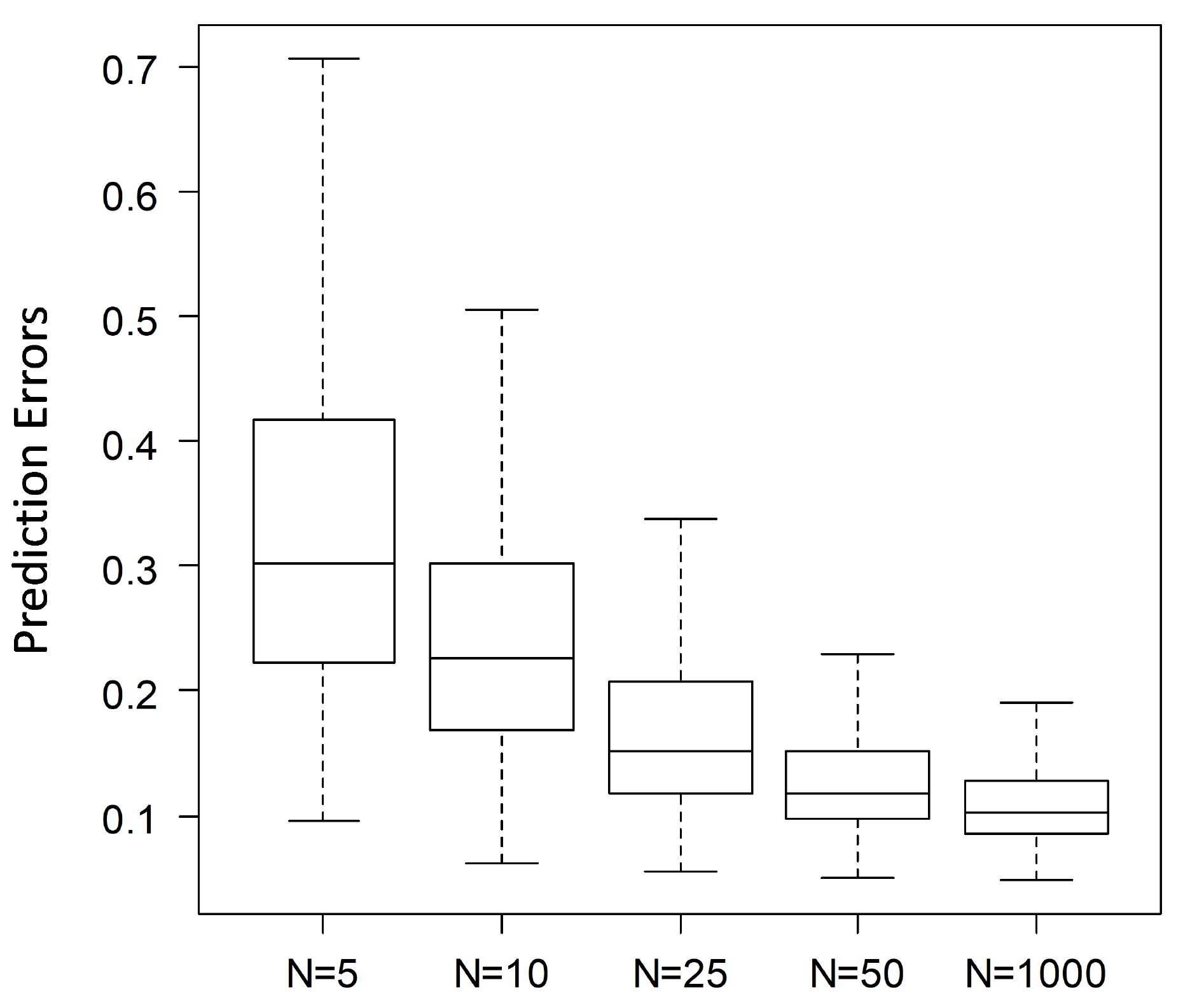}
  \caption{PoE results with different $N$}
  \label{fig:pic9}
\end{figure}

In this setting, we aim to compare the performance of the MGCP-RD in a case with a moderate number of outputs ($N=50$) and no negative transfer. Also, we also aim to illustrate the importance of MGP models in extrapolation which has not been fully exploited in literature. We adopt the quadratic function example from \cite{han2009prediction} with some modifications. The multivariate output model for the $N$ curves are generated according to $y_i(x)=1+e^{II}x^2+\epsilon_i(x)$ for $i \in \mathcal{I}$ and $e^{II}\sim uniform(0.8,1.2)$. The number of observations for the $N-1$ training output is $p=20$ evenly spaced points for $x \in [0,10]$, while for the $N$th function to be predicted we generated $p=10$ evenly spaced points for $x \in [0,7]$.  The test points are evenly spaced across $[0,10]$ and measurement noise standard deviation is set to $\sigma=1$ for all outputs. Due to the long model building time, only $W=50$ iterations for the MGCP and MGCP-I are conducted. For the MGCP-I, all $p=20$ design points in $[0,10]$ are used as  inducing variables. Also, in this setting we implement the $\ell_1$ penalization. The importance of this case is that we are able to test all benchmarked models in a rather high dimensional parameter space with higher computational complexity.  The results are shown in Figure \ref{fig:pic8}, while an illustrative example of the MGCP-RD results in shown in \ref{fig:pic9}.

Based on the results we can obtain some important insights. First, from a computational perspective, and as shown in Figure \ref{fig:pic8}, the model building time of the MGCP and MGCP-I is extremely prohibitive. It takes on average 13 hours to build one MGCP model and this is with a one dimensional input and a moderate number of outputs ($N=50$). The mean building time for each iteration was half a second for the GCP, 6.8 minutes for the MGCP-RD and 104 minutes for the MGCP-P. Although the MGCP-P considers paired models however to predict the $N$th output we need to fit $N(N-1)/2$ pairwise submodels, unlike our method where only $N-1$ models need to be built. These results illustrate why non-separable MGP models are only used in low dimensional settings. As mentioned previously, for the MGCP-RD, we parallelized computational only over 2 systems (25 models sequentially fit in each system), however, with more computational power, the building time for the MGCP-RD can be significantly reduced. 

Second, in addition to the severe computational drawback of the MGCP, its predictive accuracy greatly degrades in such a high dimensional space. This is intuitively understandable, as minimizing the negative likelihood in such a high dimensional ($4950$) search space is a prohibitive task for any search algorithm. The MLE will directly get trapped in a local minima and will not be able to move even if different starting points are tried, therefore leading to suboptimal parameter estimates with undesirable properties. Besides that, the computation is plagued by numerical issues associated with inverting the $990\times990$ covariance matrix at each iteration of the search algorithm. Similar results have also been shown in \citet{li2016pairwise,li2018pairwise,shi2011gaussian}, where MGP models tend to loose accuracy in a  high dimensional parameter space. This issue is also faced in the MGCP-I, which is not able to address the large parameter space challenge, despite tackling the computational complexity where it only requires the inversion of $20 \times 20$ covariance matrices. It is important to note that multivariate statistical modeling often encounters functional data with $N>>50$ \citep{mcfarland2008calibration,ramsay2006functional}, thus the  aforementioned drawbacks significantly increase in severity with more outputs.

Third, in Figure \ref{fig:pic9} below we compare our the MGCP-RD results in cases with different output number $N$ under setting II specifications.  As shown in the figure, as we increase $N$ the prediction errors significantly decrease. This results further highlights the efficiency of the weighted PoE model specifically when $N$ is large.

\begin{figure}[H]
\centering
\makebox[\textwidth]{\includegraphics[keepaspectratio=true,width=\textwidth]{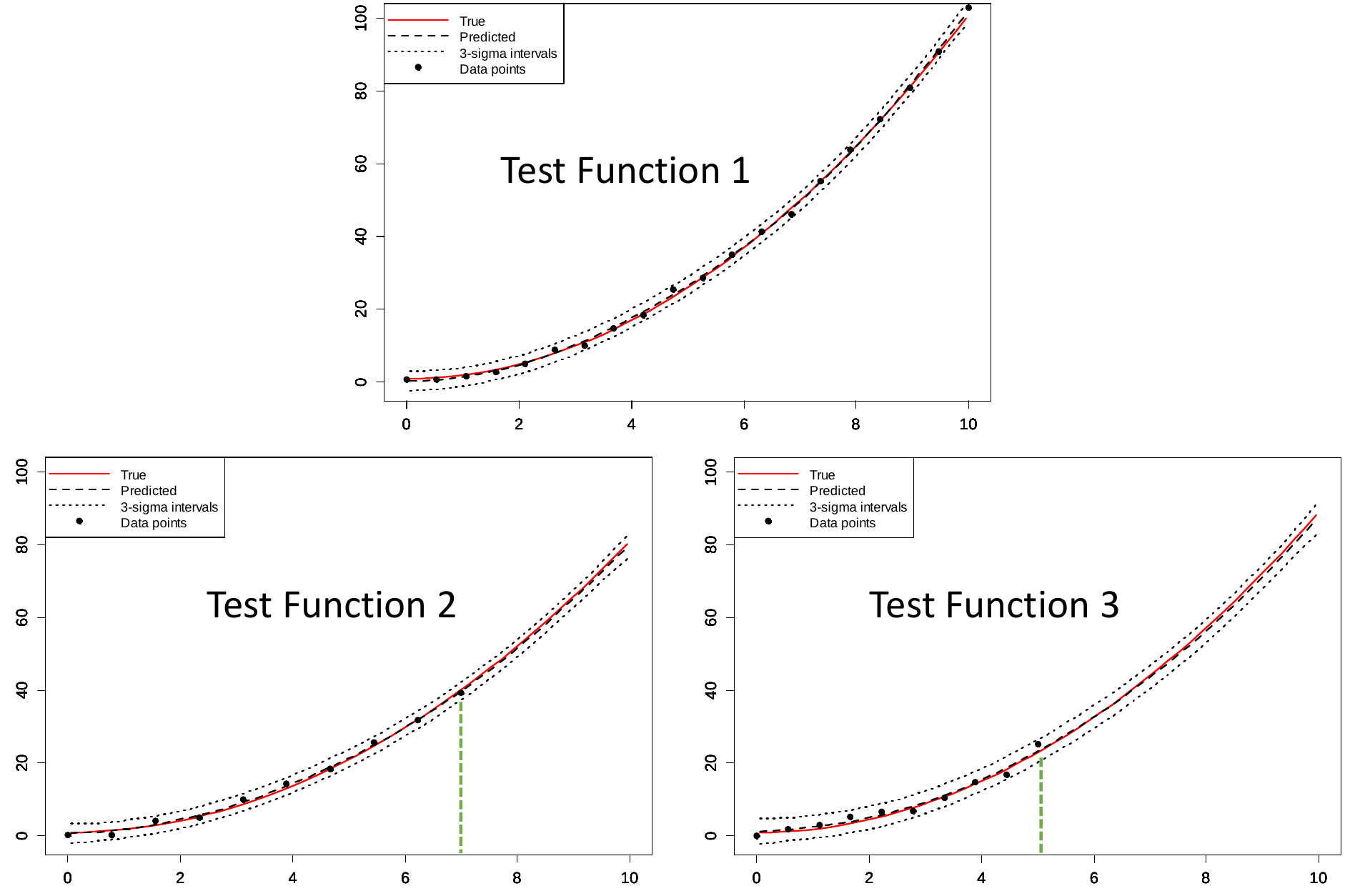}}
  \caption{Intrapolation vs extrapolation}
  \label{fig:pic10}
\end{figure} 

Finally, in Figure \ref{eq:raed8} we do not report the GCP prediction accuracy to maintain scale, since the mean prediction error is 21.78. The reason for this poor performance is related to the fact the GP in traditional settings cannot extrapolate as correlation goes to zero for data points that are far away from the observed data. However, the MGP is able to borrow strength from other observed outputs to predict the future evolution of a specific output. In other words, extrapolation in the MGP can be seen as interpolation across different output. To highlight this aspect, Figure \ref{fig:pic10} illustrates the MGCP-RD performance in extrapolation.

In Figure \ref{fig:pic10}, for function 1 we perform interpolation, while for functions 2 and 3 we extrapolate following setting II specifications. However, in function 3, we generate $p=10$ evenly spaced points for $x \in [0,5]$ instead of $x \in [0,7]$. The figure clearly shows that one main advantage of MGP models is the ability to extrapolate an output when other correlated outputs are observed over a larger domain. This advantage of MGP models indeed has many practical applications in cases where extrapolation might be also of interest to the user.

\subsubsection{Setting III}
In this setting, we aim to compare the performance of the MGCP-RD when negative transfer of knowledge exists. We establish a simple setting with few number of outputs ($N=8$). Motivated by $M/M/1$ queuing systems, the multivariate output for the $N$ functions is generated according to:

\begin{itemize}
\item $i \in \{1,2,3,4\}$:   $y^{(1)}_i(x)=x^2+\epsilon_i(x)$ for $x \in [0,0.8]$
\item $i \in \{5,6\}$:   $y^{(2)}_i(x)=x^2/(2(1-x))+\epsilon_i(x)$ for $x \in [0,0.8]$ 
\item $i \in \{7,8\}$:   $y^{(3)}_i(x)=x^2/(1-x)+\epsilon_i(x)$ for $x \in [0,0.8]$
\end{itemize}

\begin{figure}[H]
\centering
\includegraphics[keepaspectratio=true,scale=0.4]{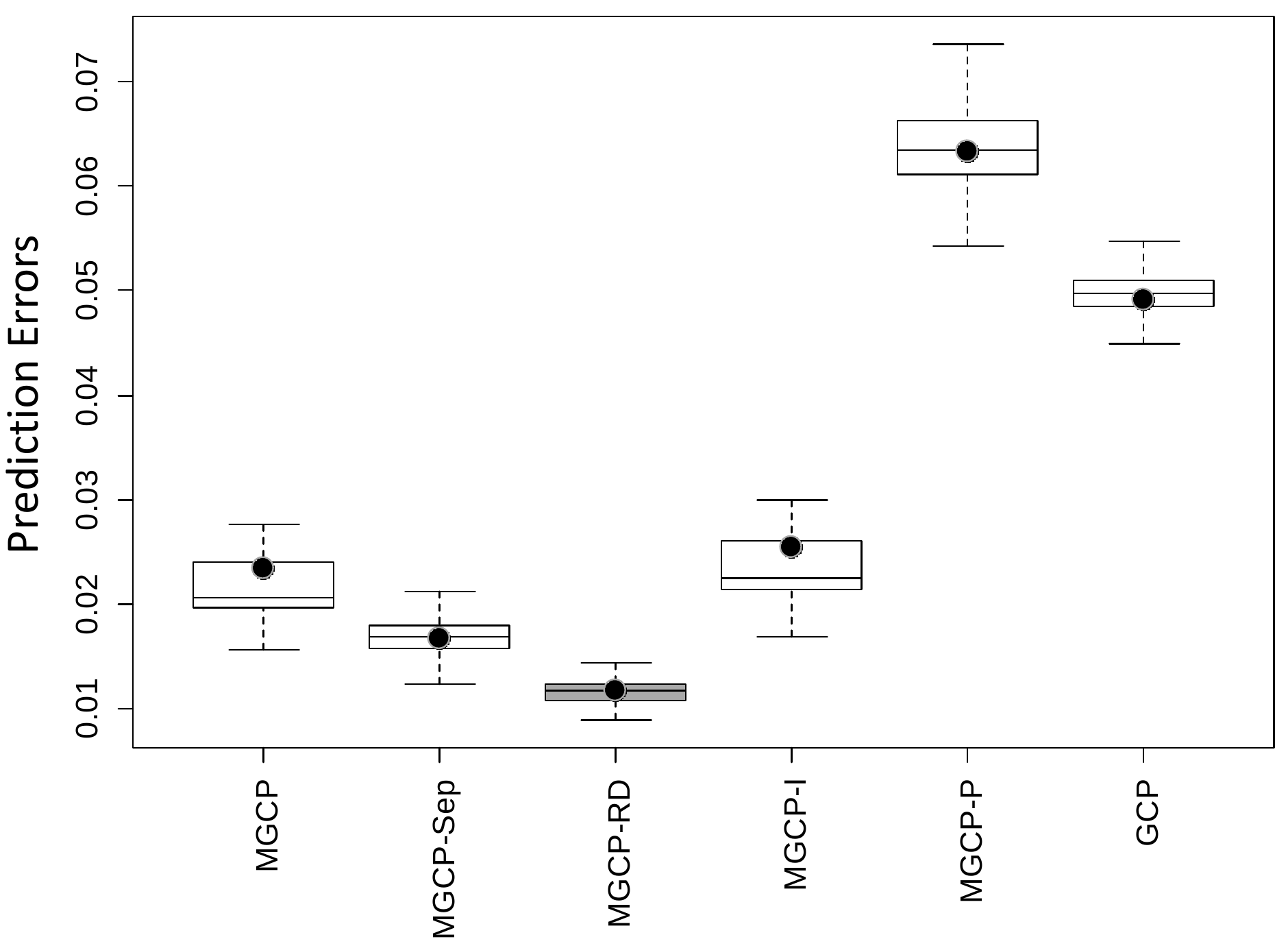}
  \caption{Setting III results}
  \label{fig:pic11}
\end{figure}

The number of observations per signal is $p=7$ evenly spaced points for $x \in [0,0.8]$, the test points are evenly spaced across $[0,0.8]$ and measurement noise standard deviation is set to $\sigma=0.005$ for all outputs. In this setting, if we assume $x$ to be the system utilization, then $y^{(2)}_i(x)$ and $y^{(3)}_i(x)$ respectively define the steady state closed form equations of the expected Queue time and Queue length in an $M/M/1$ system, where the inter-arrival time is exponentially distributed with rate 2 \citep{kleinrock1976queueing}. All $p=7$ design points in $[0,0.8]$ are used as inducing variables for the MGCP-I and we implement the $\ell_1$ penalization for the MGCP-RD. Following our general settings, for each iteration we find the MAE of the $N$th function to be predicted which belongs to $y^{(3)}_i(x)$ and represents the expected queue length. Also, we benchmark with one other method denoted as MGCP-Sep. In MGCP-Sep, the MGCP is used to predict the $N$th output using only the training signals with the same functional form, i.e. we fit outputs $i \in \{7,8\}$ separately using the MGCP. Note that, the importance of our Model Setting III, is that the we are able to test all benchmarked models in a setting where outputs behave according to different functional forms. The results are shown in Figure \ref{fig:pic11}.

The results in Figure \ref{fig:pic11} clearly illustrate the ability of our model to minimize negative transfer while borrowing strength from other correlated output. As shown in the figure, MGCP-Sep outperformed MGCP. This confirms that negative transfer is occurring since if outputs belonging to $y^{(3)}_i(x)$ are analyzed separately the results are better than the full model (MGCP). More interestingly, we have that MGCP-RD outperformed MGCP-Sep. The reason is that $y^{(3)}_i(x)$ and $y^{(2)}_i(x)$ are highly correlated, therefore, MGCP-RD was able to learn this cross correlation of  $y^{(3)}_i(x)$ with $y^{(2)}_i(x)$ while at the same time avoiding negative transfer with $y^{(1)}_i(x)$. However, MGCP-Sep is not able to learn from the cross correlation between $y^{(3)}_i(x)$ and $y^{(2)}_i(x)$. Indeed, in this model setting we observe that $\hat{\xi}_0=0$ for pairwise models including $y^{(3)}_i(x)$ and $y^{(1)}_i(x)$  which indicates that these outputs possess no common features and should be predicted independently.  Finally, one important observation is that MGCP-P behaved worse than GCP. As previously mentioned, averaging parameter estimates is specifically dangerous in cases with different functional forms as parameter estimates from different submodels will greatly fluctuate.

\subsubsection{Setting IV}
\begin{figure}[H]
\centering
\includegraphics[keepaspectratio=true, scale=0.6]{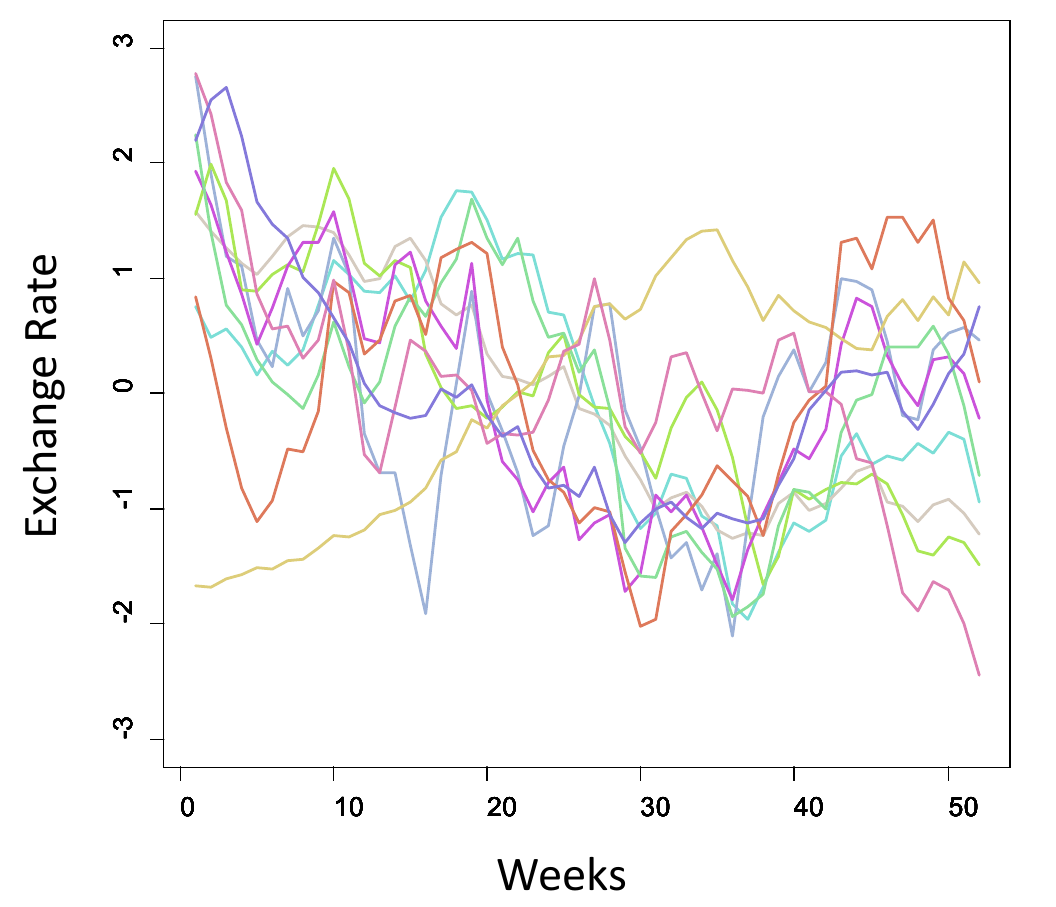}
  \caption{Data illustration}
  \label{fig:pic12}
\end{figure}

In this setting, the goal is to predict the foreign exchange rate compared to the United States dollar currency. The data comes from the pacific exchange rate service  (http://\allowbreak fx.sauder.ubc.ca/data.html). Our analysis utilized the exchange rates of the top ten international currencies (Canadian Dollar CAD/USD, Euro EUR/USD, Japanese Yen JPY/USD,	Great British Pound GBP/USD, Swiss Franc CHF/USD, Australian Dollar AUD/USD, Hong Kong Dollar HKD/USD, New Zealand Dollar NZD/USD, South Koreon Won KRW/ USD, Mexican Peso MXN/USD) during the 52 weeks of the 2017 calender year. The data is illustrated in Figure \ref{fig:pic12}. Each output is adjusted to have zero mean and unit variance. We use a leave-one-out cross validation approach to evaluate the performance of the MGCP-RD and the benchmarked methods. We iteratively treat one exchange currency rate as the test output and the remaining 9 currencies as the training set. This procedure is repeated for the 10 currencies. For each test output we randomly remove 13 data points (25$\%$ of the data from a specific output) and test the model capability to recover the true underlying values at these test points. For the MGCP, all remaining input points are used as inducing variables for the MGCP-I and we implement the $\ell_1$ penalization for the MGCP-RD.

\begin{figure}[H]
\centering
\makebox[\textwidth]{\includegraphics[keepaspectratio=true,scale=0.7]{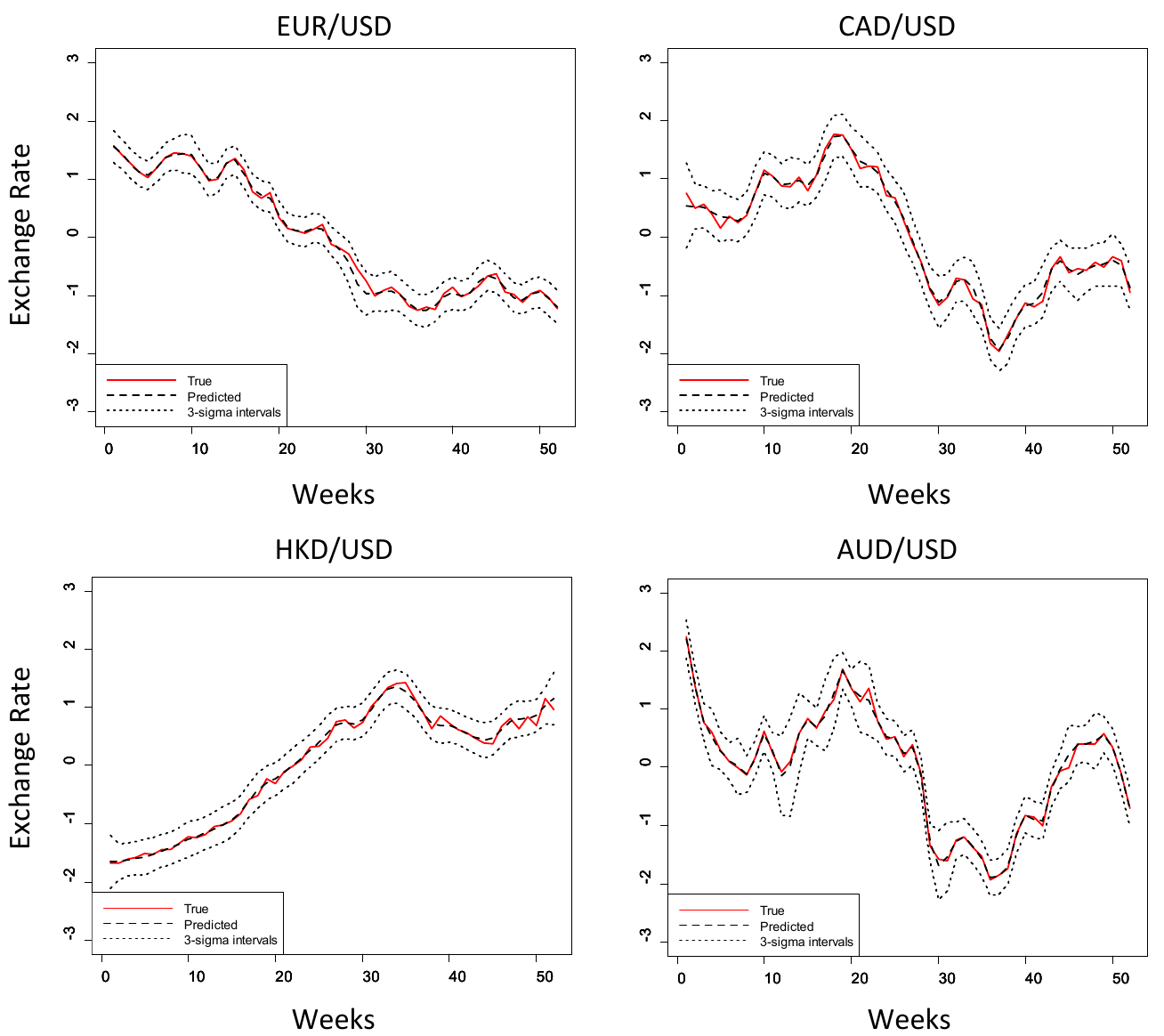}}
  \caption{MGCP-RD predictive results}
  \label{fig:pic13}
\end{figure}

First we provide some illustrative results in Figure \ref{fig:pic13}. As shown in the figure, the MGCP-RD was able to able to efficiently recover the underlying truth. For performance accuracy comparison we also use the standardized mean square error (SMSE) defined in \citep{rasmussen2004gaussian}. The results in  in Table \ref{table:1} show that the MGCP-RD was significantly able to outperform the benchmarked methods. This result in intuitively understandable as  based on Figure \ref{fig:pic12}, the trends display clear heterogeneity and thus negative transfer is a key issue when integratively modeling the exchange rates.

\vspace{-1.5em}

\begin{table}[H]
\begin{center}
 \begin{tabular}{||c c c c c c||} 
 \hline
  & MGCP & MGCP-RD & MGCP-I & MGCP-P & GCP \\ [0.5ex] 
 \hline\hline
 MAE & 0.203 & 0.156 & 0.218 & 0.952 & 0.247 \\ 
 std.error & (0.160) & (0.068) & (0.236) & (0.544) & (0.182) \\
 \hline
 SMSE & 1.666 & 1.293 & 1.788 & 3.261 & 1.703 \\
 std.error & (1.783) & (0.861) & (2.250) & (6.002) & (1.935) \\[1ex] 
 \hline
\end{tabular}
 \caption{Setting results}
\label{table:1}
\end{center}
\end{table}

\section{Conclusion}

MGP models established using a CP construction offer a general and flexible solution for multiple output regression. Despite that, the added flexibility arises serious computational challenges even with a moderate number of outputs due to the significant increase in computational demands and number of parameters to be estimated. Further, the integrative analysis of multiple outputs implicitly assumes that these outputs share some commonalities. However, if this does not hold, negative transfer of knowledge may occur. In this paper, we try to simultaneously address the computational (computational complexity, high dimensional parameter space) and negative transfer challenges. To do so, we propose a regularized pairwise modeling approach for MGCP models that has excellent scalability and minimizes the negative transfer of knowledge between uncorrelated outputs. The proposed approach is based on distributing MGCP estimation through bivariate GP submodels which are individually estimated. Predictions are then made through combining predictions from the bivariate models within a Bayesian framework. Interestingly, pairwise modeling turns out to possess unique characteristics which allows to tackle the challenge of negative transfer through penalizing shared latent functions. The modeling framework is generic in terms of the choice of the kernel function and can scale to arbitrarily large datasets by parallelization. We also provide statistical guarantees for the proposed method, extent our method to separable molding cases and demonstrate its advantageous features through numerical studies.  The numerical studies illustrate that we can (1) achieve similar prediction performance as the full multivariate approach when the output dimension is low, (2) outperform the full multivariate approach, with only a fraction of its computational needs, when the output dimension is high, (3) outperform the full multivariate approach when some functions are uncorrelated even when the output dimension is low.

One important extension of this model lies in the domain of functional graphical models. In such models, nodes are functions rather than random variables. In fact, and since a GP itself is an undirected graphical model, the MGP represents a fully connected undirected functional graphical model where each node represents an output. In our pairwise approach, we are encouraging independence between pairs of functions through our regularization framework. However, an interesting extension would be to extend this regularization framework, to build conditional independence amongst functions in a similar sense to Lemma 1. The main challenge however remains in providing a mapping between the covariance matrix and the precision matrix which control conditional independence between the outputs. We will work along this line and report the results in the future.

\newpage

\acks{We would like to acknowledge support for this project
from the National Science Foundation (NSF grant \#1561512)}.

\appendix
\section*{Appendix A.}
In this appendix we prove the following Lemmas from Section 3.4:\\

\noindent
{\bf Lemma 1}.  (\emph{Multivariate}) {\it Given that $pr(\bm{y}|\bm{X},\bm{\theta})=\mathcal{N}(\bm{0}_P,\bm{\Omega}^{-1})$, then $\bm{\Omega}_{ij}=\bm{0}$  if and only if the multivariate Gaussian random vectors $\bm{y}_i$ and $\bm{y}_j$ are conditionally independent, i.e. $\mbox{cov}(y_i^c,y_j^{c^{\prime}}|\ddot{\bm{y}})=0$ for every $c \in \{1,..,p_i\}$ and $c^{\prime} \in \{1,..,p_j\}$}.\\

\noindent
{\bf Lemma 2}. (\emph{Bivariate}) {\it Given that $pr(\bm{y}_i,\bm{y}_j|\bm{X}_1,\bm{X}_2,\bm{\theta}^{\prime})= \mathcal{N} \begin{pmatrix} \bm{0}_{p_i+p_j}, &\begin{pmatrix}
\bm{\Omega}_{ii} & \bm{\Omega}_{ij}\\
\bm{\Omega}_{ij}^t & \bm{\Omega}_{jj} 
\end{pmatrix}^{-1} \end{pmatrix}$, then  $\bm{\Omega}_{ij}=\bm{0}$ if and only if, the multivariate Gaussian random vectors $\bm{y}_i$ and $\bm{y}_j$ are independent, i.e. $\mbox{cov}(y_i^c,y_j^{c^{\prime}})=0$ for every $c \in \{1,..,p_i\}$ and $c^{\prime} \in \{1,..,p_j\}$}.    \hfill\BlackBox\\

\noindent
{\bf Proof}. We first prove Lemma 2 and deduce Lemma 1 accordingly. Let $\bm{C}_{{\bm y}_{i},{\bm y}_{j}}$ denote the covariance between the random vectors $\bm{y}_i$ and $\bm{y}_j$, where $\begin{pmatrix}
\bm{\Omega}_{ii} & \bm{\Omega}_{ij}\\
\bm{\Omega}_{ij}^t & \bm{\Omega}_{jj} 
\end{pmatrix} = \begin{pmatrix} 
\bm{C}_{\bm{y}_i,\bm{y}_i} & \bm{C}_{\bm{y}_i,\bm{y}_j}\\
\bm{C}_{\bm{y}_i,\bm{y}_j}^t & \bm{C}_{\bm{y}_j,\bm{y}_j} 
 \end{pmatrix}^{-1}=\mbox{cov}(\bm{y}_i,\bm{y}_j)^{-1}$. Using the inverse variance Lemma, we have that

\begin{equation}
\label{eq:raed18}
\begin{pmatrix}
\bm{\Omega}_{ii} & \bm{\Omega}_{ij}\\
\bm{\Omega}_{ij}^t & \bm{\Omega}_{jj} 
\end{pmatrix}= 
\begin{pmatrix}
\bm{C}_{\bm{y}_i,\bm{y}_i}^{-1}+ \bm{H}_{2}^t \mbox{cov}(\bm{y}_j|\bm{y}_i)^{-1} \bm{H}_{2}   & -\bm{H}_{2}^t \mbox{cov}(\bm{y}_j|\bm{y}_i)^{-1}\\
-\mbox{cov}(\bm{y}_j|\bm{y}_i)^{-1} \bm{H}_{2} & \mbox{cov}(\bm{y}_j|\bm{y}_i)^{-1} 
\end{pmatrix} ,
\end{equation}

\noindent
where $\bm{H}_{2}=\bm{C}_{\bm{y}_i,\bm{y}_j}^t \bm{C}_{\bm{y}_i,\bm{y}_i}^{-1}$ and $\mbox{cov}(\bm{y}_j|\bm{y}_i)=\bm{C}_{\bm{y}_j,\bm{y}_j}-\bm{C}_{\bm{y}_i,\bm{y}_j}^t \bm{C}_{\bm{y}_i,\bm{y}_i}^{-1} \bm{C}_{\bm{y}_i,\bm{y}_j}$. We now have that $\bm{\Omega}_{ij}=-\bm{H}_{2}^t \mbox{cov}(\bm{y}_j|\bm{y}_i)^{-1}=-\bm{C}_{\bm{y}_i,\bm{y}_i}^{-1}\bm{C}_{\bm{y}_i,\bm{y}_j}\mbox{cov}(\bm{y}_j|\bm{y}_i)^{-1}$. However, since  $\mbox{cov}(\bm{y}_i,\bm{y}_j)^{-1}$ is positive definite then both $\bm{C}_{\bm{y}_i,\bm{y}_i}$ and $\mbox{cov}(\bm{y}_j|\bm{y}_i)$ are positive definite and thus $\bm{\Omega}_{ij}=\bm{0}$ if and only if $\bm{C}_{\bm{y}_i,\bm{y}_j} = \bm{0}$.

To prove Lemma 1, we consider the random vectors $\bm{y}_i$, $\bm{y}_j$ and $\ddot{\bm{y}}$ where $\ddot{\bm{y}}=\{\bm{y}\}/ \{y_i^c,y_j^{c^{\prime}}\}$. Now replace $\bm{y}_i$ by $\ddot{\bm{y}}$ and replace $\bm{y}_j$ by the partitioned vector $[\bm{y}_i^t,\bm{y}_j^t]^t$. Following (\ref{eq:raed18}), we observe that $\bm{\Omega}_{jj} =\mbox{cov}(\bm{y}_i,\bm{y}_j|\ddot{\bm{y}})^{-1}$. By applying the inverse variance Lemma again, but this time to  $\mbox{cov}(\bm{y}_i,\bm{y}_j|\ddot{\bm{y}})^{-1}$ instead of $\mbox{cov}(\bm{y}_i,\bm{y}_j)^{-1}$ we have that $\bm{\Omega}_{jj} =\mbox{cov}(\bm{y}_i,\bm{y}_j|\ddot{\bm{y}})^{-1}=$

\begin{equation}
\label{eq:raed19} 
\begin{pmatrix}
\bm{H}_3   & -\bm{H}_{1}^t \mbox{cov}(\bm{y}_j|\bm{y}_i,\ddot{\bm{y}})^{-1}\\
-\mbox{cov}(\bm{y}_j|\bm{y}_i,\ddot{\bm{y}})^{-1} \bm{H}_{1} & \mbox{cov}(\bm{y}_j|\bm{y}_i,\ddot{\bm{y}})^{-1} 
\end{pmatrix} ,
\end{equation}

\noindent
where $\bm{H}_3=\mbox{cov}({\bm{y}_i,\bm{y}_i}|\ddot{\bm{y}})^{-1}+ \bm{H}_{1}^t \mbox{cov}(\bm{y}_j|\bm{y}_i,\ddot{\bm{y}})^{-1} \bm{H}_{1}$ and $\bm{H}_{1}=\mbox{cov}({\bm{y}_i,\bm{y}_j|\ddot{\bm{y}}})^t \mbox{cov}({\bm{y}_i,\bm{y}_i|\ddot{\bm{y}}})^{-1}$. Then following Lemma 2, the off-diagonal block is zero in this case, if and only if the multivariate Gaussian random vectors $\bm{y}_i$ and $\bm{y}_j$ are conditionally independent where $\mbox{cov}(y_i^c,y_j^{c^{\prime}}|\ddot{\bm{y}})=0$ for every $c \in \{1,..,p_i\}$ and $c^{\prime} \in \{1,..,p_j\}$.

\section*{Appendix B.}
In this appendix we expand on ${\partial  \bm{C}_{ij} }/{\partial \theta_{ij}^{(n)}}$ from Section 4.1. Recall that $\theta_{ij}^{(n)} \in \bm{\theta}_{ij}^t=\{\bm{\theta}_{f_{ij}}^t,\bm{\sigma}_{ij}^t\}^t$, $\bm{C}_{ij}=\bm{C}_{{\bm f}_{ij},{\bm f}_{ij}}+ \bm{\Sigma}_{ij}$ and $\bm{\Lambda}_{qi}$ is a $D \times D$ positive definite diagonal matrix allowing different length scales for each dimension. For instance if $D =2$ then $\bm{\Lambda}_{qi}=\begin{pmatrix}
\nu_{qi(1)}^2 & 0\\
0 & \nu_{qi(2)}^2
\end{pmatrix}$. We first expand on ${\partial  \bm{C}_{ii} }/{\partial \theta_{ij}^{(n)}}$, where, as shown in Section 4.1, the covariance of the marginal process is $\mbox{cov}^f_{ii}(\bm x,{\bm x}^{\prime})=\sum_{q=\{0,i\}} \xi_q^2 \alpha_{qi}^2 \mbox{exp}( -\frac{1}{4} \bm{d}^t \bm{\Lambda}_{qi} \bm{d} )=\sum_{q=\{0,i\}} \xi_q^2 \alpha_{qi}^2 \mbox{exp}\big( -\frac{1}{4} \sum_{c=1}^D d_{(c)}^2 \nu_{0i(c)}^2\big)$ for $c\in\{1,...,D\}$.

\begin{gather*}
\frac{\partial \mbox{cov}^f_{ii}}{\partial \xi_0^2 }=2\xi_0 \alpha_{0i}^2 \mbox{exp}( -\frac{1}{4} \bm{d}^t \bm{\Lambda}_{0i} \bm{d} ); \ \frac{\partial \mbox{cov}^f_{ii}}{\partial \alpha_{0i}^2 }=2\xi^2_0 \alpha_{0i} \mbox{exp}( -\frac{1}{4} \bm{d}^t \bm{\Lambda}_{0i} \bm{d} );  
\\  \frac{\partial \mbox{cov}^f_{ii}}{\partial \nu_{0i(c)}^2 }=-\frac{1}{2} \xi^2_0 \alpha_{0i}^2 d_{(c)}^2 \nu_{0i(c)}^2 \mbox{exp}( -\frac{1}{4} \bm{d}^t \bm{\Lambda}_{0i} \bm{d} ); \ \frac{\partial \mbox{cov}^f_{ii}}{\partial \sigma_{i} }=2\sigma_i \tau_{ij}\tau_{x,x^{\prime}} . 
\end{gather*}

We exclude $\frac{\partial \mbox{cov}^f_{ii}}{\partial \xi_i^2 }$, $\frac{\partial \mbox{cov}^f_{ii}}{\partial \alpha_{ii}^2 }$ and  $\frac{\partial \mbox{cov}^f_{ii}}{\partial \nu_{ii(c)}^2 }$ due to similarity with their counterparts above. Now when $i \neq j$, we have that $\mbox{cov}^f_{ij}(\bm x,{\bm x}^{\prime})=\xi_0^2 \tilde{\omega}^0_{ij} \mbox{exp}( -\frac{1}{2} \bm{d}^t {\bm{\Phi}^{0}_{ij}}^{-1} \bm{d} )$ where $\tilde{\omega}^0_{ij}= 2^\frac{D}{2}\alpha_{0i} \alpha_{0j} \allowbreak |\bm{\Lambda}_{0i}|^{\frac{1}{4}}|\bm{\Lambda}_{0j}|^{\frac{1}{4}}/|\bm{\Lambda}_{0i}+\bm{\Lambda}_{0j}|^\frac{1}{2}$, and ${\bm{\Phi}^{q}_{ij}}^{-1}=(\bm{\Lambda}_{qi}^{-1}+\bm{\Lambda}_{qj}^{-1})^{-1}$. Let $c^{\prime}=\{1,...,D\}/\{c\}$ then

\begin{gather*}
\frac{\partial \mbox{cov}^f_{ij}}{\partial \xi_0^2 }=2\xi_0  \tilde{\omega}^0_{ij}  \mbox{exp}( -\frac{1}{2} \bm{d}^t {\bm{\Phi}^{0}_{ij}}^{-1} \bm{d} ); \ \frac{\partial \mbox{cov}^f_{ij}}{\partial \alpha_{0i}^2 }=\xi_0^2  2^\frac{D}{2}\alpha_{0j} \frac{|\bm{\Lambda}_{0i}|^{\frac{1}{4}}|\bm{\Lambda}_{0j}|^{\frac{1}{4}}}{|\bm{\Lambda}_{0i}+\bm{\Lambda}_{0j}|^\frac{1}{2}}  \mbox{exp}( -\frac{1}{2} \bm{d}^t {\bm{\Phi}^{0}_{ij}}^{-1} \bm{d} ); \\
\frac{\partial \mbox{cov}^f_{ij}}{\partial \nu_{0i(c)}^2 }=\bar{B} \xi_0^2 2^\frac{D}{2}\alpha_{0i} \alpha_{0j} \mbox{exp}( -\frac{1}{2} \bm{d}^t {\bm{\Phi}^{0}_{ij}}^{-1} \bm{d} )+ \xi_0^2  2^\frac{D}{2}\alpha_{0i}\alpha_{0j} \frac{|\bm{\Lambda}_{0i}|^{\frac{1}{4}}|\bm{\Lambda}_{0j}|^{\frac{1}{4}}}{|\bm{\Lambda}_{0i}+\bm{\Lambda}_{0j}|^\frac{1}{2}} \bar{\bar B};\\
 \bar{\bar B}=-d_{(c)}^2 \frac{ \nu_{0i(c)}\nu^4_{0j(c)} } {(\nu^2_{0i(c)} + \nu^2_{0j(c)})^2}; \bar{B}= \\ 
\frac{\nu_{0i(c)}|\bm{\Lambda}_{0j}|^{\frac{1}{4}} \Big(\frac{1}{2} \prod\limits_{c^{\prime}} \nu^2_{0i(c^{\prime})} |\bm{\Lambda}_{0i}|^{-\frac{3}{4}}  |\bm{\Lambda}_{0i}+\bm{\Lambda}_{0j}|^\frac{1}{2} - |\bm{\Lambda}_{0i}|^{\frac{1}{4}}\nu_{0i(c)} |\bm{\Lambda}_{0i}+\bm{\Lambda}_{0j}|^{-\frac{1}{2}}   \prod\limits_{c^{\prime}} (\nu^2_{0i(c^{\prime})} + \nu^2_{0j(c^{\prime})}) \Big)}{|\bm{\Lambda}_{0i}+\bm{\Lambda}_{0j}|} .
\end{gather*}

\noindent
We also exclude $\frac{\partial \mbox{cov}^f_{ij}}{\partial \alpha_{0j}^2 }$ and $\frac{\partial \mbox{cov}^f_{ij}}{\partial \nu_{0j(c)}^2 }$ due to similarity with their counterparts above.

\section*{Appendix C.}
In this appendix we prove the following theorem from Section 4.2:\\

\noindent
\textbf{Theorem 1.} \emph{Suppose that $\xi_0=0$, then the predictive distribution of the bivariate model at any new input $\bm{x}_0 \in \mathcal{X}$ reduces to that of a univariate model where $pr(y_i(\bm{x}_0)|\bm{y}_{ij})=pr(y_i(\bm{x}_0)|\bm{y}_{i})$ and $pr(y_j(\bm{x}_0)|\bm{y}_{ij})=pr(y_j(\bm{x}_0)|\bm{y}_{j})$ such that
$$
\begin{cases} 
pr(y_i(\bm{x}_0)|\bm{y}_{ij})=\mathcal{N} \Big(    \bm{C}^t_{\bm{f}_i,f_i^0} \bm{\Omega}_{ii} \bm{y}_i,\ C_{f_i^0,f_i^0}+ \sigma^2_i-\bm{C}^t_{\bm{f}_i,f_i^0} \bm{\Omega}_{ii} \bm{C}_{\bm{f}_i,f_i^0} \Big) \\
pr(y_j(\bm{x}_0)|\bm{y}_{ij})=\mathcal{N} \Big(    \bm{C}^t_{\bm{f}_j,f_j^0} \bm{\Omega}_{jj}\bm{y}_j,\ C_{f_j^0,f_j^0}+ \sigma^2_j-\bm{C}^t_{\bm{f}_j,f_j^0} \bm{\Omega}_{jj} \bm{C}_{\bm{f}_j,f_j^0} \Big)
\end{cases}$$
where for $c\in\{i,j\}$, $\bm{\Omega}_{cc}=(\bm{C}_{\bm{f}_c,\bm{f}_c} + \sigma^2_c \bm{I}_{p_c})^{-1}\in\mathcal{R}^{p_c \times p_c}$, $\bm{C}_{\bm{f}_c,f_c^0}=[\mbox{cov}^f_{cc}(\bm{x}_0,\bm{x}_{c1}),...,\allowbreak \mbox{cov}^f_{cc}(\bm{x}_0,\bm{x}_{cp_c})]^t$,  $C_{f_c^0,f_c^0}=\mbox{cov}^f_{cc}(\bm{x}_0,{\bm{x}_0})$ and $\mbox{cov}^f_{cc}(\bm x,{\bm x}^{\prime})= \xi_c^2 \int_{-\infty}^{\infty} K_{cc}(\bm{u})K_{cc}(\bm{u}-\bm{d}) d\bm{u}$}. \hfill\BlackBox\\

\noindent
{\bf Proof}.  Based on (\ref{eq:raed9}), $\xi_0=0$ implies that, for $i \neq j$,  $\mbox{cov}^f_{ij}(\bm x,{\bm x}^{\prime})=\xi_0^2 \int_{-\infty}^{\infty} K_{0i}(\bm{u})K_{0j}(\bm{u}-\bm{d}) d\bm{u} =0$ for every $\bm{d}$. Therefore, we have that $\bm{C}_{ij}=
\bm{C}_{{\bm f}_{ij},{\bm f}_{ij}}+ \bm{\Sigma}_{ij}= 
\begin{pmatrix} 
\bm{C}_{\bm{f}_i,\bm{f}_i} & \bm{0}_{p_i\times p_j}\\
\bm{0}_{p_i\times p_j} & \bm{C}_{\bm{f}_j,\bm{f}_j} 
 \end{pmatrix}+
 \begin{pmatrix} 
\sigma^2_i \bm{I}_{p_i} & \bm{0}\\
\bm{0} & \sigma^2_j \bm{I}_{p_j} 
 \end{pmatrix}$.  Applying (\ref{eq:raed3}) under a bivariate setting we we have that

\begin{equation}
  \label{eq:raed20}
pr(y_i(\bm{x}_0)|\bm{y}_{ij})=\mathcal{N} \Big(    \bm{C}^t_{{\bm f}_{ij},f_i^0} \bm{C}_{ij}^{-1}\bm{y}_{ij},\ C_{f_i^0,f_i^0}+ \sigma^2_i-\bm{C}^t_{{\bm f}_{ij},f_i^0} \bm{C}_{ij}^{-1} \bm{C}_{{\bm f}_{ij},f_i^0} \Big).
\end{equation}

Recall, $\bm{C}_{{\bm f}_{ij},f_i^0}=[\bm{C}^t_{\bm{f}_i,f_i^0},\bm{C}^t_{\bm{f}_j,f_i^0}]^t$ where $\bm{C}^t_{\bm{f}_i,f_i^0}=[\mbox{cov}^f_{ii}(\bm{x}_0,\bm{x}_{i1}),...,\mbox{cov}^f_{ii}(\bm{x}_0,\bm{x}_{ip_i})]^t$  and 
$\bm{C}_{\bm{f}_j,f_i^0}=[\mbox{cov}^f_{ij}(\bm{x}_0,\bm{x}_{j1}),...,\mbox{cov}^f_{ij}(\bm{x}_0,\bm{x}_{jp_j})]^t$. However, since $\mbox{cov}^f_{ij}(\bm x,{\bm x}^{\prime})=0$ for $i \neq j$ then $\bm{C}_{\bm{f}_j,f_i^0}=\bm{0}^t_{p_j}$, therefore

\begin{gather*}
\bm{C}^t_{{\bm f}_{ij},f_i^0} \bm{C}_{ij}^{-1}\bm{y}_{ij}=
[\bm{C}^t_{\bm{f}_i,f_i^0},\bm{0}_{p_j}] \begin{pmatrix} 
(\bm{C}_{\bm{f}_i,\bm{f}_i} + \sigma^2_i \bm{I}_{p_i})^{-1} & \bm{0}_{p_i\times p_j}\\
\bm{0}_{p_i\times p_j} & (\bm{C}_{\bm{f}_j,\bm{f}_j} + \sigma^2_j \bm{I}_{p_j})^{-1}
 \end{pmatrix}[\bm{y}_i^t,\bm{y}_j^t]^t
 \\=\bm{C}^t_{\bm{f}_i,f_i^0} \bm{\Omega}_{ii} \bm{y}_i
\end{gather*}
\begin{gather*}
\bm{C}^t_{{\bm f}_{ij},f_i^0} \bm{C}_{ij}^{-1} \bm{C}_{{\bm f}_{ij},f_i^0}=
[\bm{C}^t_{\bm{f}_i,f_i^0},\bm{0}_{p_j}] \begin{pmatrix} 
(\bm{C}_{\bm{f}_i,\bm{f}_i} + \sigma^2_i \bm{I}_{p_i})^{-1} & \bm{0}_{p_i\times p_j}\\
\bm{0}_{p_i\times p_j} & (\bm{C}_{\bm{f}_j,\bm{f}_j} + \sigma^2_j \bm{I}_{p_j})^{-1}
 \end{pmatrix}[\bm{C}^t_{\bm{f}_i,f_i^0},\bm{0}_{p_j}]^t\\
 =\bm{C}^t_{\bm{f}_i,f_i^0} \bm{\Omega}_{ii} \bm{C}_{\bm{f}_i,f_i^0}
\end{gather*}

\noindent
Similarly, we can obtain the proof $pr(y_j(\bm{x}_0)|\bm{y}_{ij})$. 

\section*{Appendix D.}
In this appendix we prove the following theorems from Section 4.2:\\

\noindent
\textbf{Theorem 2.} \emph{If $z_2=\mbox{max}\{|\mathbb{P}^{\prime\prime}_\lambda(|\xi_0^{\ast}|)|:\xi_0^{\ast} \neq 0\} \rightarrow 0$, then there exits a local minimzer $\hat{\bm{\theta}}_{ij}$ for $\ell_\mathbb{P}(\bm{\theta}_{ij})$, such that $||\hat{\bm{\theta}}_{ij}-\bm{\theta}^{\ast}_{ij}||=O(r^{-1}_{\mathrm{p}}+z_1)$, where $z_1=\mbox{max}\{\mathbb{P}^{\prime}_\lambda(|\xi_0^{\ast}|):\xi_0^{\ast} \neq 0\}$   }. \\

\noindent
\textbf{Theorem 3.} \emph{Assume that $\xi_0^{\ast}=0$ and the parameters $\ddot{\bm{\theta}}_{ij}=\{\bm{\theta}_{ij}\}/ \{\xi_0\}$ satisfy $r_{\mathrm{p}}$ consistency in theorem 2. Then if $\underset{\mathrm{p} \rightarrow \infty}{lim \ inf} \ \underset{\xi_0 \rightarrow 0^{+}}{lim \ inf} \ \frac{1}{\lambda} \mathbb{P}^{\prime}_\lambda(\xi_0) > 0$, $\lambda \rightarrow 0$ and ${\mathrm{p}\lambda}/{r_{\mathrm{p}}} \ \rightarrow \infty$ as $\mathrm{p} \rightarrow \infty$, we have that $\underset{\mathrm{p} \rightarrow \infty}{lim} pr(\hat{\xi}_0=0)=1$}. \hfill\BlackBox\\

\noindent
{\bf Proof}. First we note that for the the negative log-likelihood $(\ell^{\prime})$ and penalty function $(\mathbb{P}^{\prime}_\lambda(|\xi_0|))$ the \raisebox{0.3ex}{``$\prime$"} notation on a function implies a derivative. 
As previously mentioned the MLE for the unpenalized likelihood  $\ell(\bm{\theta}_{ij})$ is $r_{\mathrm{p}}$ consistent where $r_{\mathrm{p}}$ is a sequence such that $r_{\mathrm{p}} \rightarrow \infty$ as $\mathrm{p} \rightarrow \infty$. Therefore, we have that $r_{\mathrm{p}}^{-1} \ell^{\prime}(\bm{\theta}_{ij})=O(1)$ and  $||\hat{\bm{\theta}}_{ij}-\bm{\theta}^{\ast}_{ij}||=O(r^{-1}_{\mathrm{p}})$ \citep{basawa1980statistical,basawa1976asymptotic}. This result is a direct extension of the well known root-$\mathrm{p}$ consistency of the MLE based on independent and identically distributed normal observations, which holds under the usual regularity conditions (please refer to chapter 7 of \cite{basawa1980statistical}). In theorems 2 and 3 we aim to study the asymptotic properties of the penalized likelihood $\ell_\mathbb{P}(\bm{\theta}_{ij})=\ell(\bm{\theta}_{ij})+\mathbb{P}_\lambda(|\xi_0|) \ $. The proofs provide similar results as in \cite{fan2001variable}, but defined within our model specifications and based on dependent observations. To be consistent with \cite{fan2001variable} notation, instead of minimizing the negative log-likelihood we maximize the log-likelihood whose form follows $\ell_{\mathbb{P}+}(\bm{\theta}_{ij})=-\ell_\mathbb{P}(\bm{\theta}_{ij})=-\ell(\bm{\theta}_{ij})-\mathbb{P}_\lambda(|\xi_0|)=\ell_+(\bm{\theta}_{ij})-\mathbb{P}_\lambda(|\xi_0|)$. Also we follow their convention by multiplying by $\mathrm{p}$ the penalty function, i.e. $\ell_{\mathbb{P}+}(\bm{\theta}_{ij})= \ell_+(\bm{\theta}_{ij})-\mathrm{p}\mathbb{P}_\lambda(|\xi_0|)$. To prove theorem 2, we need to show that for any given $\varepsilon>0$ there exists a large constant $\mathcal{G}$ such that

\begin{equation}
\label{eq:raed21}
pr \big( \ \underset{||\bm{g}||=\mathcal{G}}{\mbox{sup}} \ \ell_{\mathbb{P}+}(\bm{\theta}^{\ast}_{ij}+\rho \bm{g}) < \ell_{\mathbb{P}+}(\bm{\theta}^{\ast}_{ij}) \ \big)\geq 1-\varepsilon \ ,
\end{equation}

\noindent
where $\rho=r^{-1}_{\mathrm{p}}+z_1$. This equation implies that, with a probability at least $1-\varepsilon$, there exists a local maximum in the ball $\{ \bm{\theta}^{\ast}_{ij}+\rho \bm{g}: ||\bm{g}|| \leq \mathcal{G} \}$, where the local maximizer $\hat{\bm{\theta}}_{ij}$ satisfies  $||\hat{\bm{\theta}}_{ij}-\bm{\theta}^{\ast}_{ij}||=O(\rho)=O(r^{-1}_{\mathrm{p}}+z_1)$. Expanding on $\ell_\mathbb{P}(\bm{\theta}^{\ast}_{ij}+\rho \bm{g}) - \ell_\mathbb{P}(\bm{\theta}^{\ast}_{ij})$, we have that
\begin{equation}
\label{eq:raed22}
\begin{split}
 \ell_{\mathbb{P}+}(\bm{\theta}^{\ast}_{ij}+\rho \bm{g}) - \ell_{\mathbb{P}+}(\bm{\theta}^{\ast}_{ij}) & =
\big[ \ell_+(\bm{\theta}^{\ast}_{ij}+\rho \bm{g})- \ell_+(\bm{\theta}^{\ast}_{ij}) \big] - \mathrm{p} \big[ \mathbb{P}_\lambda(|\xi^{\ast}_0+\rho g_{\xi_0}|) - \mathbb{P}_\lambda(|\xi^{\ast}_0|) \big] ,
\end{split}
\end{equation}

\noindent
where $g_{\xi_0}$ denotes the element in $\bm{g}$ corresponding to $\xi_0$. Under the assumption that $\mathbb{P}_\lambda(|\xi_0|)\geq 0 $ and $\mathbb{P}_\lambda(0)= 0$, and if $\xi^{\ast}_0=0$ then  $\ell_{\mathbb{P}+}(\bm{\theta}^{\ast}_{ij}+\rho \bm{g}) - \ell_{\mathbb{P}+}(\bm{\theta}^{\ast}_{ij}) \leq
\ell_+(\bm{\theta}^{\ast}_{ij}+\rho \bm{g})- \ell_+(\bm{\theta}^{\ast}_{ij})$ as $\mathbb{P}_\lambda(|\xi^{\ast}_0+g_{\xi_0}|) - \mathbb{P}_\lambda(|\xi^{\ast}_0|) = \mathbb{P}_\lambda(|\rho g_{\xi_0}|)  \geq 0$. Using a Taylor expansion we have that $\ell_+(\bm{\theta}^{\ast}_{ij}+\rho \bm{g})=\ell_+(\bm{\theta}^{\ast}_{ij}) + \rho {\ell_+^{\prime}}(\bm{\theta}^{\ast}_{ij})^t \bm{g}-\frac{\rho^2}{2}\bm{g}^t \mathbb{I}(\bm{\theta}^{\ast}_{ij})\bm{g}\{1+o(1)\}$ where ${\ell_+^{\prime}}(\bm{\theta}^{\ast}_{ij})^t$ is the gradient vector of $\ell_+$ evaluated at $\bm{\theta}^{\ast}_{ij}$ and $\mathbb{I}$ is a finite positive definite information matrix at $\bm{\theta}^{\ast}_{ij}$. Therefore, through applying a Taylor expansion also on $\mathbb{P}_\lambda(|\xi^{\ast}_0+\rho g_{\xi_0}|)$, we have that 
\begin{equation}
\label{eq:raed23}
\begin{split}
 \ell_{\mathbb{P}+}(\bm{\theta}^{\ast}_{ij}+\rho \bm{g}) - \ell_{\mathbb{P}+}(\bm{\theta}^{\ast}_{ij})  & \leq
\big[ \rho {\ell_+^{\prime}}(\bm{\theta}^{\ast}_{ij})^t \bm{g}-\frac{\rho^2}{2}\bm{g}^t \mathbb{I}(\bm{\theta}^{\ast}_{ij})\bm{g}\{1+o(1)\} \big]
 \\&- \mathrm{p}\big[ \rho \mathbb{P}^{\prime}_\lambda(|\xi_0^{\ast}|) \mbox{sign}(\xi_0^{\ast})g_{\xi_0} + \rho^2 \mathbb{P}^{\prime\prime}_\lambda(|\xi_0^{\ast}|)g_{\xi_0}^2\{1+o(1)\} \big] .
\end{split}
\end{equation}

Note that $r_{\mathrm{p}}^{-1} \ell^{\prime}(\bm{\theta}_{ij})=O(1)$, and the penalty form is similar to that of \cite{fan2001variable}, thus the rest of the proof is identical to theorem 1 in \cite{fan2001variable}. Regarding theorem 3, we need to show that $\underset{\mathrm{p} \rightarrow \infty}{lim} pr(\hat{\xi}_0=0)=1$ if $\xi_0^{\ast}=0$. First we have that
\begin{equation}
\label{eq:raed24}
\begin{split}
\frac{\partial \ell_{\mathbb{P}+}(\bm{\theta}_{ij})}{\partial \xi_0 } =   \frac{\partial \ell_+(\bm{\theta}_{ij})}{\partial \xi_0 } - \mathrm{p}\mathbb{P}^{\prime}_\lambda(|\xi_0|) \mbox{sign}(\xi_0) 
\end{split} .
\end{equation}

\noindent
Following a first order Taylor expansion, (\ref{eq:raed24}) can be written as
\begin{equation}
\label{eq:raed25}
\begin{split}
\frac{\partial \ell_{\mathbb{P}+}(\hat{\bm{\theta}}_{ij})}{\partial \xi_0 } &=   \frac{\partial \ell_+(\bm{\theta}^{\ast}_{ij}) }{\partial \xi_0 } + \sum_n \frac{\partial^2 \ell_{+}(\bm{\theta}^{\ast}_{ij})}{\partial \xi_0 \partial \theta_{ij}^{(n)}  } (\hat{\bm{\theta}}_{ij}^{(n)}-\bm{\theta}^{\ast (n)}_{ij})    
\\& + \sum_n \sum_{n^{\prime}}\frac{\partial^3 \ell_{+}(\bm{\theta}^{\ast\ast}_{ij})}{\partial \xi_0 \partial \theta_{ij}^{(n)} \theta_{ij}^{(n^{\prime})}  } (\hat{\bm{\theta}}_{ij}^{(n)}-\bm{\theta}^{\ast (n)}_{ij})(\hat{\bm{\theta}}_{ij}^{(n^{\prime})}-\bm{\theta}^{\ast (n^{\prime})}_{ij})          - 
\mathrm{p} \mathbb{P}^{\prime}_\lambda(|\hat{\xi}_0|) \mbox{sign}(\hat{\xi}_0) ,
\end{split}
\end{equation}

\noindent
for some $\bm{\theta}^{\ast\ast}_{ij} \in (\bm{\theta}^{\ast}_{ij},\hat{\bm{\theta}}_{ij})$. However, since  $\ell_{+}(\hat{\bm{\theta}}_{ij})$ is of order $O(r_{\mathrm{p}})$ and given that $||\hat{\bm{\theta}}_{ij}-\bm{\theta}^{\ast}_{ij}||=O(r^{-1}_{\mathrm{p}})$ we have $\frac{\partial \ell_{\mathbb{P}+}(\hat{\bm{\theta}}_{ij})}{\partial \xi_0 }= \lambda\mathrm{p}\big[  -\frac{1}{\lambda}\mathbb{P}^{\prime}_\lambda(|\hat{\xi}_0|) \mbox{sign}(\hat{\xi}_0) + O\Big(\frac{r_{\mathrm{p}}}{\lambda\mathrm{p}} \Big) \big]$. From here and since, $\underset{\mathrm{p} \rightarrow \infty}{lim \ inf} \ \underset{\xi_0 \rightarrow 0^{+}}{lim \ inf} \ \frac{1}{\lambda} \mathbb{P}^{\prime}_\lambda(\xi_0) > 0$ and ${\mathrm{p}\lambda}/{r_{\mathrm{p}}} \ \rightarrow \infty$, then the sign of $\frac{\partial \ell_{\mathbb{P}+}(\hat{\bm{\theta}}_{ij})}{\partial \xi_0 }$ is only determined by the sign of $\hat{\xi}_0$. Thus, $\frac{\partial \ell_{\mathbb{P}+}(\hat{\bm{\theta}}_{ij})}{\partial \xi_0 } < 0 $ for $0< \hat{\xi}_0 < \varepsilon$ and $\frac{\partial \ell_{\mathbb{P}+}(\hat{\bm{\theta}}_{ij})}{\partial \xi_0 } > 0 $ for $-\varepsilon< \hat{\xi}_0 < 0 $ for some small $\varepsilon$, which is a sufficient condition to prove theorem 3.

\vskip 0.2in
\bibliography{raedcite}

\begin{thebibliography}{59}
\providecommand{\natexlab}[1]{#1}
\providecommand{\url}[1]{\texttt{#1}}
\expandafter\ifx\csname urlstyle\endcsname\relax
  \providecommand{\doi}[1]{doi: #1}\else
  \providecommand{\doi}{doi: \begingroup \urlstyle{rm}\Url}\fi

\bibitem[Alvarez and Lawrence(2009)]{alvarez2009sparse}
Mauricio Alvarez and Neil~D Lawrence.
\newblock Sparse convolved gaussian processes for multi-output regression.
\newblock In \emph{Advances in neural information processing systems}, pages
  57--64, 2009.

\bibitem[{\'A}lvarez et~al.(2010){\'A}lvarez, Luengo, Titsias, and
  Lawrence]{alvarez2010efficient}
Mauricio {\'A}lvarez, David Luengo, Michalis Titsias, and Neil Lawrence.
\newblock Efficient multioutput gaussian processes through variational inducing
  kernels.
\newblock In \emph{Proceedings of the Thirteenth International Conference on
  Artificial Intelligence and Statistics}, pages 25--32, 2010.

\bibitem[{\'A}lvarez and Lawrence(2011)]{alvarez2011computationally}
Mauricio~A {\'A}lvarez and Neil~D Lawrence.
\newblock Computationally efficient convolved multiple output gaussian
  processes.
\newblock \emph{Journal of Machine Learning Research}, 12\penalty0
  (May):\penalty0 1459--1500, 2011.

\bibitem[Alvarez et~al.(2012)Alvarez, Rosasco, Lawrence,
  et~al.]{alvarez2012kernels}
Mauricio~A Alvarez, Lorenzo Rosasco, Neil~D Lawrence, et~al.
\newblock Kernels for vector-valued functions: A review.
\newblock \emph{Foundations and Trends{\textregistered} in Machine Learning},
  4\penalty0 (3):\penalty0 195--266, 2012.

\bibitem[Barry et~al.(1996)Barry, Jay, and Hoef]{barry1996blackbox}
Ronald~Paul Barry, M~Jay, and Ver Hoef.
\newblock Blackbox kriging: spatial prediction without specifying variogram
  models.
\newblock \emph{Journal of Agricultural, Biological, and Environmental
  Statistics}, pages 297--322, 1996.

\bibitem[Basawa(1980)]{basawa1980statistical}
Ishwar~V Basawa.
\newblock \emph{Statistical Inferences for Stochasic Processes: Theory and
  Methods}.
\newblock Elsevier, 1980.

\bibitem[Basawa et~al.(1976)Basawa, Feigin, and Heyde]{basawa1976asymptotic}
IV~Basawa, PD~Feigin, and CC~Heyde.
\newblock Asymptotic properties of maximum likelihood estimators for stochastic
  processes.
\newblock \emph{Sankhy{\=a}: The Indian Journal of Statistics, Series A}, pages
  259--270, 1976.

\bibitem[Boyle(2007)]{boyle2007gaussian}
Phillip Boyle.
\newblock Gaussian processes for regression and optimisation.
\newblock 2007.

\bibitem[Boyle and Frean(2005)]{boyle2005dependent}
Phillip Boyle and Marcus Frean.
\newblock Dependent gaussian processes.
\newblock In \emph{Advances in neural information processing systems}, pages
  217--224, 2005.

\bibitem[Calder and Cressie(2007)]{calder2007some}
Catherine~A Calder and Noel Cressie.
\newblock Some topics in convolution-based spatial modeling.
\newblock \emph{Proceedings of the 56th Session of the International Statistics
  Institute}, pages 22--29, 2007.

\bibitem[Cao and Fleet(2014)]{cao2014generalized}
Yanshuai Cao and David~J Fleet.
\newblock Generalized product of experts for automatic and principled fusion of
  gaussian process predictions.
\newblock \emph{arXiv preprint arXiv:1410.7827}, 2014.

\bibitem[Caruana(1998)]{caruana1998multitask}
Rich Caruana.
\newblock Multitask learning.
\newblock In \emph{Learning to learn}, pages 95--133. Springer, 1998.

\bibitem[Colosimo et~al.(2014)Colosimo, Cicorella, Pacella, and
  Blaco]{colosimo2014profile}
Bianca~M Colosimo, Paolo Cicorella, Massimo Pacella, and Marzia Blaco.
\newblock From profile to surface monitoring: Spc for cylindrical surfaces via
  gaussian processes.
\newblock \emph{Journal of Quality Technology}, 46\penalty0 (2):\penalty0
  95--113, 2014.

\bibitem[Conti and O’Hagan(2010)]{conti2010Bayesian}
Stefano Conti and Anthony O’Hagan.
\newblock Bayesian emulation of complex multi-output and dynamic computer
  models.
\newblock \emph{Journal of statistical planning and inference}, 140\penalty0
  (3):\penalty0 640--651, 2010.

\bibitem[Conti et~al.(2009)Conti, Gosling, Oakley, and
  O'Hagan]{conti2009gaussian}
Stefano Conti, John~Paul Gosling, Jeremy~E Oakley, and Anthony O'Hagan.
\newblock Gaussian process emulation of dynamic computer codes.
\newblock \emph{Biometrika}, 96\penalty0 (3):\penalty0 663--676, 2009.

\bibitem[Deisenroth and Ng(2015)]{deisenroth2015distributed}
Marc~Peter Deisenroth and Jun~Wei Ng.
\newblock Distributed gaussian processes.
\newblock \emph{arXiv preprint arXiv:1502.02843}, 2015.

\bibitem[Donoho and Johnstone(1995)]{donoho1995adapting}
David~L Donoho and Iain~M Johnstone.
\newblock Adapting to unknown smoothness via wavelet shrinkage.
\newblock \emph{Journal of the american statistical association}, 90\penalty0
  (432):\penalty0 1200--1224, 1995.

\bibitem[Fan and Li(2001)]{fan2001variable}
Jianqing Fan and Runze Li.
\newblock Variable selection via nonconcave penalized likelihood and its oracle
  properties.
\newblock \emph{Journal of the American statistical Association}, 96\penalty0
  (456):\penalty0 1348--1360, 2001.

\bibitem[Fieuws and Verbeke(2006)]{fieuws2006pairwise}
Steffen Fieuws and Geert Verbeke.
\newblock Pairwise fitting of mixed models for the joint modeling of
  multivariate longitudinal profiles.
\newblock \emph{Biometrics}, 62\penalty0 (2):\penalty0 424--431, 2006.

\bibitem[Fricker et~al.(2013)Fricker, Oakley, and
  Urban]{fricker2013multivariate}
Thomas~E Fricker, Jeremy~E Oakley, and Nathan~M Urban.
\newblock Multivariate gaussian process emulators with nonseparable covariance
  structures.
\newblock \emph{Technometrics}, 55\penalty0 (1):\penalty0 47--56, 2013.

\bibitem[Friedman et~al.(2001)Friedman, Hastie, and
  Tibshirani]{friedman2001elements}
Jerome Friedman, Trevor Hastie, and Robert Tibshirani.
\newblock \emph{The elements of statistical learning}, volume~1.
\newblock Springer series in statistics New York, 2001.

\bibitem[Goulard and Voltz(1992)]{goulard1992linear}
Michel Goulard and Marc Voltz.
\newblock Linear coregionalization model: tools for estimation and choice of
  cross-variogram matrix.
\newblock \emph{Mathematical Geology}, 24\penalty0 (3):\penalty0 269--286,
  1992.

\bibitem[Han et~al.(2009)Han, Santner, Notz, and Bartel]{han2009prediction}
Gang Han, Thomas~J Santner, William~I Notz, and Donald~L Bartel.
\newblock Prediction for computer experiments having quantitative and
  qualitative input variables.
\newblock \emph{Technometrics}, 51\penalty0 (3):\penalty0 278--288, 2009.

\bibitem[Haykin(2008)]{haykin2008communication}
Simon Haykin.
\newblock \emph{Communication systems}.
\newblock John Wiley \& Sons, 2008.

\bibitem[Helterbrand and Cressie(1994)]{helterbrand1994universal}
Jeffrey~D Helterbrand and Noel Cressie.
\newblock Universal cokriging under intrinsic coregionalization.
\newblock \emph{Mathematical Geology}, 26\penalty0 (2):\penalty0 205--226,
  1994.

\bibitem[Heskes(1998)]{heskes1998selecting}
Tom Heskes.
\newblock Selecting weighting factors in logarithmic opinion pools.
\newblock In \emph{Advances in neural information processing systems}, pages
  266--272, 1998.

\bibitem[Higdon(2002)]{higdon2002space}
Dave Higdon.
\newblock Space and space-time modeling using process convolutions.
\newblock In \emph{Quantitative methods for current environmental issues},
  pages 37--56. Springer, 2002.

\bibitem[Higdon et~al.(2008)Higdon, Gattiker, Williams, and
  Rightley]{higdon2008computer}
Dave Higdon, James Gattiker, Brian Williams, and Maria Rightley.
\newblock Computer model calibration using high-dimensional output.
\newblock \emph{Journal of the American Statistical Association}, 103\penalty0
  (482):\penalty0 570--583, 2008.

\bibitem[Kleinrock(1976)]{kleinrock1976queueing}
Leonard Kleinrock.
\newblock \emph{Queueing systems, volume 2: Computer applications}, volume~66.
\newblock wiley New York, 1976.

\bibitem[Kontar et~al.(2017{\natexlab{a}})Kontar, Zhou, Sankavaram, Du, and
  Zhang]{kontar2017nonparametrica}
Raed Kontar, Shiyu Zhou, Chaitanya Sankavaram, Xinyu Du, and Yilu Zhang.
\newblock Nonparametric-condition-based remaining useful life prediction
  incorporating external factors.
\newblock \emph{IEEE Transactions on Reliability}, 2017{\natexlab{a}}.

\bibitem[Kontar et~al.(2017{\natexlab{b}})Kontar, Zhou, Sankavaram, Du, and
  Zhang]{kontar2017nonparametricb}
Raed Kontar, Shiyu Zhou, Chaitanya Sankavaram, Xinyu Du, and Yilu Zhang.
\newblock Nonparametric modeling and prognosis of condition monitoring signals
  using multivariate gaussian convolution processes.
\newblock \emph{Technometrics}, \penalty0 (just-accepted), 2017{\natexlab{b}}.

\bibitem[Li and Zhou(2016)]{li2016pairwise}
Yongxiang Li and Qiang Zhou.
\newblock Pairwise meta-modeling of multivariate output computer models using
  nonseparable covariance function.
\newblock \emph{Technometrics}, 58\penalty0 (4):\penalty0 483--494, 2016.

\bibitem[Li et~al.(2018)Li, Zhou, Huang, and Zeng]{li2018pairwise}
Yongxiang Li, Qiang Zhou, Xiaohu Huang, and Li~Zeng.
\newblock Pairwise estimation of multivariate gaussian process models with
  replicated observations: Application to multivariate profile monitoring.
\newblock \emph{Technometrics}, 60\penalty0 (1):\penalty0 70--78, 2018.

\bibitem[Majumdar and Gelfand(2007)]{majumdar2007multivariate}
Anandamayee Majumdar and Alan~E Gelfand.
\newblock Multivariate spatial modeling for geostatistical data using convolved
  covariance functions.
\newblock \emph{Mathematical Geology}, 39\penalty0 (2):\penalty0 225--245,
  2007.

\bibitem[Mardia and Watkins(1989)]{mardia1989multimodality}
KV~Mardia and AJ~Watkins.
\newblock On multimodality of the likelihood in the spatial linear model.
\newblock \emph{Biometrika}, 76\penalty0 (2):\penalty0 289--295, 1989.

\bibitem[McFarland et~al.(2008)McFarland, Mahadevan, Romero, and
  Swiler]{mcfarland2008calibration}
John McFarland, Sankaran Mahadevan, Vicente Romero, and Laura Swiler.
\newblock Calibration and uncertainty analysis for computer simulations with
  multivariate output.
\newblock \emph{AIAA journal}, 46\penalty0 (5):\penalty0 1253--1265, 2008.

\bibitem[Melkumyan and Ramos(2011)]{melkumyan2011multi}
Arman Melkumyan and Fabio Ramos.
\newblock Multi-kernel gaussian processes.
\newblock In \emph{IJCAI Proceedings-International Joint Conference on
  Artificial Intelligence}, volume~22, page 1408, 2011.

\bibitem[Ng and Deisenroth(2014)]{ng2014hierarchical}
Jun~Wei Ng and Marc~Peter Deisenroth.
\newblock Hierarchical mixture-of-experts model for large-scale gaussian
  process regression.
\newblock \emph{arXiv preprint arXiv:1412.3078}, 2014.

\bibitem[Nguyen et~al.(2014)Nguyen, Bonilla, et~al.]{nguyen2014collaborative}
Trung~V Nguyen, Edwin~V Bonilla, et~al.
\newblock Collaborative multi-output gaussian processes.
\newblock In \emph{UAI}, pages 643--652, 2014.

\bibitem[Osborne et~al.(2008)Osborne, Roberts, Rogers, Ramchurn, and
  Jennings]{osborne2008towards}
Michael~A Osborne, Stephen~J Roberts, Alex Rogers, Sarvapali~D Ramchurn, and
  Nicholas~R Jennings.
\newblock Towards real-time information processing of sensor network data using
  computationally efficient multi-output gaussian processes.
\newblock In \emph{Proceedings of the 7th international conference on
  Information processing in sensor networks}, pages 109--120. IEEE Computer
  Society, 2008.

\bibitem[Paciorek and Schervish(2004)]{paciorek2004nonstationary}
Christopher~J Paciorek and Mark~J Schervish.
\newblock Nonstationary covariance functions for gaussian process regression.
\newblock In \emph{Advances in neural information processing systems}, pages
  273--280, 2004.

\bibitem[Pan and Yang(2010)]{pan2010survey}
Sinno~Jialin Pan and Qiang Yang.
\newblock A survey on transfer learning.
\newblock \emph{IEEE Transactions on knowledge and data engineering},
  22\penalty0 (10):\penalty0 1345--1359, 2010.

\bibitem[Qian et~al.(2008)Qian, Wu, and Wu]{qian2008gaussian}
Peter Z~G Qian, Huaiqing Wu, and CF~Jeff Wu.
\newblock Gaussian process models for computer experiments with qualitative and
  quantitative factors.
\newblock \emph{Technometrics}, 50\penalty0 (3):\penalty0 383--396, 2008.

\bibitem[Qui{\~n}onero-Candela and Rasmussen(2005)]{quinonero2005unifying}
Joaquin Qui{\~n}onero-Candela and Carl~Edward Rasmussen.
\newblock A unifying view of sparse approximate gaussian process regression.
\newblock \emph{Journal of Machine Learning Research}, 6\penalty0
  (Dec):\penalty0 1939--1959, 2005.

\bibitem[Ramsay(2006)]{ramsay2006functional}
James~O Ramsay.
\newblock \emph{Functional data analysis}.
\newblock Wiley Online Library, 2006.

\bibitem[Rasmussen(2004)]{rasmussen2004gaussian}
Carl~Edward Rasmussen.
\newblock Gaussian processes in machine learning.
\newblock In \emph{Advanced lectures on machine learning}, pages 63--71.
  Springer, 2004.

\bibitem[Schwaighofer and Tresp(2003)]{schwaighofer2003transductive}
Anton Schwaighofer and Volker Tresp.
\newblock Transductive and inductive methods for approximate gaussian process
  regression.
\newblock In \emph{Advances in Neural Information Processing Systems}, pages
  977--984, 2003.

\bibitem[Shi and Choi(2011)]{shi2011gaussian}
Jian~Qing Shi and Taeryon Choi.
\newblock \emph{Gaussian process regression analysis for functional data}.
\newblock CRC Press, 2011.

\bibitem[Stein and Corsten(1991)]{stein1991universal}
A~Stein and LCA Corsten.
\newblock Universal kriging and cokriging as a regression procedure.
\newblock \emph{Biometrics}, pages 575--587, 1991.

\bibitem[Tajbakhsh et~al.(2014)Tajbakhsh, Aybat, and
  Del~Castillo]{tajbakhsh2014sparse}
Sam~Davanloo Tajbakhsh, Necdet~Serhat Aybat, and Enrique Del~Castillo.
\newblock Sparse precision matrix selection for fitting gaussian random field
  models to large data sets.
\newblock \emph{arXiv preprint arXiv:1405.5576}, 2014.

\bibitem[Thi{\'e}baux and Pedder(1987)]{thiebaux1987spatial}
H~Jean Thi{\'e}baux and MA~Pedder.
\newblock \emph{Spatial objetive analysis: with applications in atmospheric
  science}.
\newblock Number 519.24 THI. 1987.

\bibitem[Tresp(2000)]{tresp2000Bayesian}
Volker Tresp.
\newblock A bayesian committee machine.
\newblock \emph{Neural computation}, 12\penalty0 (11):\penalty0 2719--2741,
  2000.

\bibitem[Ver~Hoef and Barry(1998)]{ver1998constructing}
Jay~M Ver~Hoef and Ronald~Paul Barry.
\newblock Constructing and fitting models for cokriging and multivariable
  spatial prediction.
\newblock \emph{Journal of Statistical Planning and Inference}, 69\penalty0
  (2):\penalty0 275--294, 1998.

\bibitem[Whittaker(2009)]{whittaker2009graphical}
Joe Whittaker.
\newblock \emph{Graphical models in applied multivariate statistics}.
\newblock Wiley Publishing, 2009.

\bibitem[Wikle(2002)]{wikle2002kernel}
Christopher~K Wikle.
\newblock A kernel-based spectral model for non-gaussian spatio-temporal
  processes.
\newblock \emph{Statistical Modelling}, 2\penalty0 (4):\penalty0 299--314,
  2002.

\bibitem[Williams et~al.(2009)Williams, Klanke, Vijayakumar, and
  Chai]{williams2009multi}
Christopher Williams, Stefan Klanke, Sethu Vijayakumar, and Kian~M Chai.
\newblock Multi-task gaussian process learning of robot inverse dynamics.
\newblock In \emph{Advances in Neural Information Processing Systems}, pages
  265--272, 2009.

\bibitem[Wonnacott and Wonnacott(1990)]{wonnacott1990introductory}
Thomas~H Wonnacott and Ronald~J Wonnacott.
\newblock \emph{Introductory statistics}, volume~5.
\newblock Wiley New York, 1990.

\bibitem[Yuan et~al.(2012)Yuan, Liu, and Yan]{yuan2012visual}
Xiao-Tong Yuan, Xiaobai Liu, and Shuicheng Yan.
\newblock Visual classification with multitask joint sparse representation.
\newblock \emph{IEEE Transactions on Image Processing}, 21\penalty0
  (10):\penalty0 4349--4360, 2012.

\bibitem[Zhou et~al.(2011)Zhou, Qian, and Zhou]{zhou2011simple}
Qiang Zhou, Peter~ZG Qian, and Shiyu Zhou.
\newblock A simple approach to emulation for computer models with qualitative
  and quantitative factors.
\newblock \emph{Technometrics}, 53\penalty0 (3):\penalty0 266--273, 2011.

\end{thebibliography}

\end{document}